\documentclass[10pt,journal,compsoc]{IEEEtran}
\ifCLASSOPTIONcompsoc
  \usepackage[nocompress]{cite}
\else
  \usepackage{cite}
\fi

\usepackage{graphicx}
\usepackage{subfigure}
\usepackage{multirow}
\usepackage{amssymb}
\usepackage{amsmath}
\usepackage{booktabs}
\usepackage[ruled]{algorithm2e}
\usepackage{color}
\usepackage{ragged2e}
\usepackage{hyperref}
\usepackage{cite}

\ifCLASSINFOpdf
\else
\fi
\begin{document}
%
\title{1xN Pattern for Pruning Convolutional \\ Neural Networks}
%
%
%
%

\author{Mingbao Lin,
        Yuxin Zhang,
        Yuchao Li,
        Bohong Chen,
        Fei Chao,~\IEEEmembership{Member,~IEEE}, 
        Mengdi Wang,\\
        Shen Li,
        Yonghong Tian,~\IEEEmembership{Fellow,~IEEE}, 
        and Rongrong Ji,~\IEEEmembership{Senior Member,~IEEE}

\IEEEcompsocitemizethanks{


\IEEEcompsocthanksitem M. Lin, Y. Zhang, B. Chen and F. Chao are with the Media Analytics and Computing Laboratory, Department of Artificial Intelligence, School of Informatics, Xiamen University, Xiamen 361005, China.\protect
\IEEEcompsocthanksitem M. Lin is also with the Tencent Youtu Lab, Shanghai 200233, China.\protect
\IEEEcompsocthanksitem R. Ji (Corresponding  Author) is with the Media Analytics and Computing Laboratory, Department of Artificial Intelligence, School of Informatics, Xiamen University, Xiamen 361005, China, also with Institute of Artificial Intelligence, Xiamen University, Xiamen 361005, China (e-mail: rrji@xmu.edu.cn).\protect
\IEEEcompsocthanksitem Y. Li, M. Wang, and S. Li are with Alibaba Group.\protect
\IEEEcompsocthanksitem Y. Tian is with the School of Electronics Engineering and Computer Science, Peking University, Beijing 100871, China.
}

\thanks{Manuscript received April 19, 2005; revised August 26, 2015.}}

%
%

\markboth{IEEE TRANSACTIONS ON PATTERN ANALYSIS AND MACHINE INTELLIGENCE}
{Shell \MakeLowercase{\textit{et al.}}: Bare Demo of IEEEtran.cls for Computer Society Journals}
%



\IEEEtitleabstractindextext{%
\begin{abstract}
\justifying
Though network pruning receives popularity in reducing the complexity of convolutional neural networks (CNNs), it remains an open issue to concurrently maintain model accuracy as well as achieve significant speedups on general CPUs. In this paper, we propose a novel 1$\times$N pruning pattern to break this limitation. In particular, consecutive N output kernels with the same input channel index are grouped into one block, which serves as a basic pruning granularity of our pruning pattern. Our 1$\times$N pattern prunes these blocks considered unimportant. We also provide a workflow of filter rearrangement that first rearranges the weight matrix in the output channel dimension to derive more influential blocks for accuracy improvements and then applies similar rearrangement to the next-layer weights in the input channel dimension to ensure correct convolutional operations. Moreover, the output computation after our 1$\times$N pruning can be realized via a parallelized block-wise vectorized operation, leading to significant speedups on general CPUs. The efficacy of our pruning pattern is proved with experiments on ILSVRC-2012. For example, given the pruning rate of 50\% and N=4, our pattern obtains about 3.0\% improvements over filter pruning in the top-1 accuracy of MobileNet-V2. Meanwhile, it obtains 56.04ms inference savings on Cortex-A7 CPU over weight pruning. Our project is made available at \url{https://github.com/lmbxmu/1xN}.

\end{abstract}

}

\maketitle

\IEEEdisplaynontitleabstractindextext

%
\IEEEpeerreviewmaketitle

\IEEEraisesectionheading{\section{Introduction}\label{introduction}}
%
%
%
%
\IEEEPARstart
{C}{onvolutional} neural networks (CNNs) have substantially advanced varieties of computer vision tasks~\cite{Sun2014DeepLF,he2016deep,Kim2016CharacterAwareNL}.
Despite these tremendous success, newly developed networks tend to have more learnable parameters which also mean more floating-point operations (FLOPs). As a result, these CNNs can be rarely run on the general CPUs embedded devices with limited computation power~\cite{han2015learning}. By pruning the redundancy in CNNs, the emerging network pruning has become a broad consensus in favour of model deployment by both the academia and industries.


\begin{figure}[!t]
\begin{center}
\includegraphics[height=0.5\linewidth]{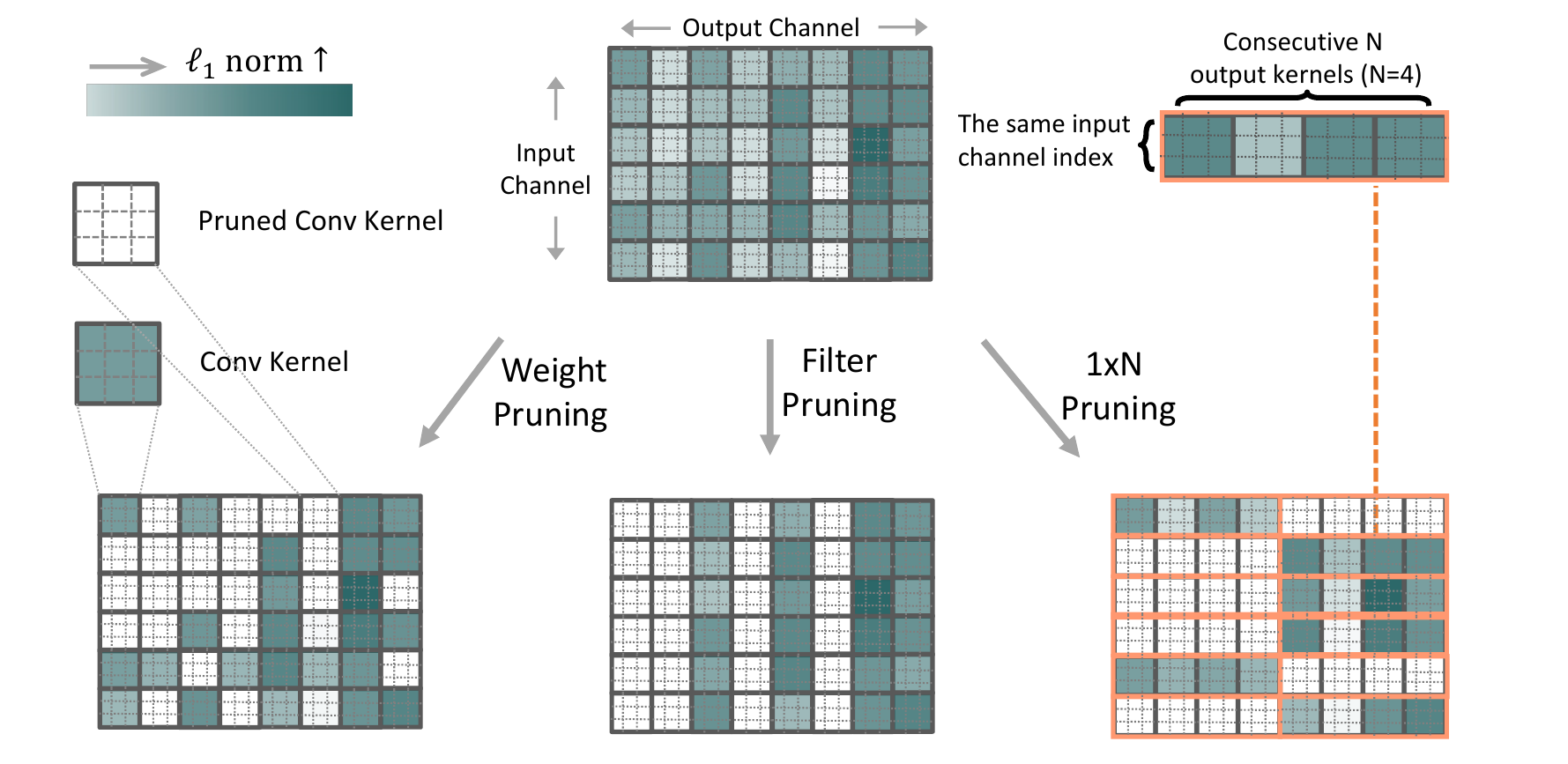}
\end{center}
\caption{\label{existing}
Comparison between existing pruning scenarios (weight pruning and filter pruning) and our 1$\times$N pruning when sparsifying convolutional weights with a shape of 8$\times$6$\times$3$\times$3 in this illustration. Given a full model, weight pruning removes some weights in the filters and filter pruning removes the whole filters. In contrast, our 1$\times$N pruning removes consecutive N output kernels with the same input channel index (N=4 in this illustration). Best viewed in colors.
}
\end{figure}

As illustrated in Fig.\,\ref{existing}, according to the basic pruning granularity, existing works accomplishing network pruning are categorized into weight pruning and filter pruning. The basic granularity of weight pruning falls into individual weights at any location of the filters or connections between full-connected layers. It essentially sparsifies the network at a fine-grained level and is demonstrated to achieve an extremely high compression rate and high accuracy performance~\cite{han2015learning,evci2020rigging,molchanov2016pruning}. However, weight pruning receives very limited speed gains since its irregular sparsity barely takes advantage of vector processing architectures such as Single Instruction Multiple Data (SIMD), and poorly utilizes memory buses. In contrast, this increases latency due to the dependent sequences of reads~\cite{choquette2021nvidia}. Recent studies~\cite{zhou2021learning,hubara2021accelerated,choquette2021nvidia,mishra2021accelerating} advocate N:M weight pruning where N out of M weights are zeros for every continuous M weights. Currently, this pattern achieves acceleration only in the case of 2:4. Besides, the acceleration is realized on the specially designed sparse Matrix Multiply-Accumulate (MMA) instructions of NVIDIA A100 towards modern single- and multi-GPU workstations, servers, clusters, and even supercomputers~\cite{choquette2021nvidia}, making it impossible to be utilized on other types of GPUs, let alone the CPUs-based platforms.

Different from weight pruning, the basic granularity of filter pruning in Fig.\,\ref{existing} consists of the whole filters. It reduces network complexity at a coarse-grained level by removing all weights in a filter. Consequently, the network structure does not change, thus the sparsified network can be well fitted by regular hardware and off-the-shelf basic linear algebra subprograms (BLAS) library to obtain acceleration. Nevertheless, filter pruning only maintains accuracy under moderate sparsity rates. Otherwise, such methods suffer more significant performance degradation than weight pruning methods. For example, a recent study~\cite{renda2020comparing} shows that a $5.96\times$ parameter reduction of ResNet-50 can well retain the accuracy performance of the original network by weight pruning, however, it is only a $1\times$ reduction in filter pruning. Though the research community has developed varieties of techniques~\cite{li2016pruning,liu2019metapruning,guo2020dmcp,lin2020hrank}, recent works~\cite{liu2018rethinking, le2021network} demonstrate that the capacity of these techniques for performance improvement is indeed limited if appropriate training settings are given to the previous works.

Above all, how to simultaneously retain the performance and achieve apparent acceleration on mobile and embedded devices becomes a challenging but valuable problem. In this paper, we propose a novel pattern of 1$\times$N pruning with its merits in realizing both high-performing accuracy and apparent CPUs acceleration for practical model deployment. Our 1$\times$N pattern provides an intermediate granular level for network pruning, which is coarser as compared to the fine-grained weight but finer as compared to the coarse-grained filter. An example of our pruning pattern that satisfies N=4 requirement is shown in Fig.\,\ref{existing}: the core distinction of our pruning from existing scenarios~\cite{lin2020hrank,liu2019metapruning,li2016pruning} lies in that our basic pruning granularity consists of consecutive N output kernels with the same input channel index. In short, we aim to remove these consecutive kernels with smaller $\ell_1$ norms that are considered less important in literature~\cite{li2016pruning}. 
Our 1$\times$N pruning in this paper follows the typical three-step pipeline of network training, pruning consecutive kernels with smaller $\ell_1$ norms, and fine-tuning the sparsified one to recover the performance.
At the point of the second step, we further propose a workflow of filter rearrangement, which rearranges the weight matrix in the output channel dimension according to the $\ell_1$ norm of each filter, based on which more influential consecutive kernels with larger $\ell_1$ norms are observed for accuracy improvements. Then, the next-layer weights are similarly rearranged in the input channel dimension to ensure the same convolutional results. In contrast to earlier developments~\cite{mao2017exploring,wortsman2019discovering,xie2019exploring} that also explore removing kernels, we have a stronger requirement of continuity on the N removed kernels, with its benefit in acceleration because these consecutive kernels can be stored continuously in the memory cache and the convolution with the inputs can proceed using a block-wise vectorized operation in parallel as analyzed in Sec.\,\ref{decoding_efficiency}.

We display multiple compression rates using the light-weight MobileNet-V1~\cite{howard2017mobilenets}, -V2~\cite{sandler2018mobilenetv2}, -V3~\cite{howard2019searching} and large-scale ResNet-50~\cite{he2016deep} on the challenging ILSVRC-2012~\cite{deng2009imagenet}, and compare our 1$\times$N pruning with weight pruning and filter pruning. The experiments suggest obvious increasing accuracy performance compared with filter pruning, and apparent inference acceleration compared with weight pruning. For example, given a pruning rate of 50\% and N=4, our 1$\times$N pattern obtains around 3.0\% improvements over filter pruning in the top-1 accuracy of MobileNet-V2 on ImageNet, meanwhile, it obtains 56.04ms inference savings on Cortex-A7 CPU compared to weight pruning which obtains no speedup.

This work addresses the problem of simultaneously maintaining accuracy and achieving general CPU speedups to enable practical model deployment on CPUs-based platforms. The key contributions of this paper include: (1) One novel pattern of 1$\times$N for network pruning. (2) A workflow of filter rearrangement for accuracy improvements. (3) Simultaneously maintaining high-performing accuracy and achieving apparent CPUs acceleration.

The remainder of this paper is organized as follows: We briefly discuss some relevant prior works in network pruning in Sec.\,\ref{realted_work}. Then, we present details of our 1$\times$N pattern for network pruning in Sec.\,\ref{methodology}. In Sec.\,\ref{experiments}, a discussion on the empirical evaluation of our method in comparison with weight pruning and filter pruning is presented. Moreover, a brief discussion on the limitation of this work is given in Sec.\,\ref{limitation}, laying out some avenues for future research in our 1$\times$N pruning pattern. We finally conclude in Sec.\,\ref{conclusion}.

\section{Related Work}\label{realted_work}

Traditional network pruning including weight pruning and filter pruning is a classical research topic. We briefly review some related works below.

\textbf{Weight Pruning}. Weight pruning dates back to Optimal Brain Damage~\cite{lecun1989optimal} and Optimal Brain Surgeon~\cite{hassibi1992second}, which prune weights based on the Hessian of the loss function. Despite their accuracy retaining, the second-order Hessian needs additional computation cost. Dong~\emph{et al}.~\cite{dong2017learning} restricted the second-order derivatives for a specific layer to enable tractable computation. 
Han \emph{et al}.~\cite{han2015learning} proposed to recursively remove small-weight connectivity and retrain the $\ell_2$-regularized subnetwork to derive smaller 
weight values. Dynamic network surgery~\cite{guo2016dynamic} performs pruning and splicing on-the-fly. The former compresses the network and the latter recovers the incorrect pruning. The lottery ticket hypothesis~\cite{frankle2019lottery} randomly initializes a dense network and trains it from scratch. The subnets with high-weight values are extracted, and retrained with the initial weight values of the original dense model. Lin \emph{et al}.~\cite{lin2020dynamic} proposed a dynamic allocation of sparsity pattern and incorporated feedback signal to reactivate prematurely pruned weights. 

\textbf{Filter Pruning}. The norm of filter weight such as $\ell_1$-norm~\cite{li2016pruning} is often considered as an indicator of filter importance. Filters with smaller norms are considered unimportant and removed. He \emph{et al}.~\cite{he2018soft} pruned the filter with $\ell_2$-norm criterion, but the pruned filters are changeable and endowed with the chance to be recovered during network training. Ding \emph{et al}.~\cite{ding2019approximated} computed the changes in the next layer’s outputs to evaluate the impact of pruning the filters. 
%
Lin \emph{et al}.~\cite{lin2020channel} used the artificial-bee-colony-based evolutionary algorithm to automatically search for the best pruning structure for each layer. He \emph{et al}.~\cite{he2018amc} leveraged reinforcement learning to sample many subnetworks from the original CNN for evaluation, and ultimately find the best compressed network. Liu \emph{et al}.~\cite{liu2019metapruning} adopted meta-learning to prune redundant filters. It trains a weight-generating meta-network in advance for subnetworks evaluation, and then searches for the best subnetwork.

\textbf{Discussion}. While a variety of approaches for network pruning have been proposed, existing methods fail to either maintain accuracy or achieve apparent speedups on the general CPUs-based platforms. Thus, it is natural for researchers to go further on pruning neural networks. This motivates our search for designing one new pruning pattern that enables general CPUs acceleration as well as maintains accuracy performance.

\section{Methodology}\label{methodology}
In this section, we introduce the intuition of our method and present its implementation details. In order to simplify the explanations, we only talk about the convolutional layers as illustrations. However, our 1$\times$N pruning can also be applied to the fully-connected layers since their weights can be regarded as 1 x 1 convolutions.

\subsection{Preliminaries}\label{preliminaries}
We start with notation definitions. Considering a pre-trained $L$-layer CNN model $F$, we denote its filter set as $\mathbf{W} = \{ \mathbf{W}^i \}_{i=1}^L$ with $\mathbf{W}^i = \{ \mathbf{W}^i_j \}_{j=1}^{n^i} \in \mathbb{R}^{n^i \times m^i \times h^i \times w^i}$, where $n^i$, $m^i$, $h^i$ and $w^i$ respectively indicate the number of output channel, input channel, kernel height and kernel width in the $i$-th layer; $\mathbf{W}^i$ is the filter set for the $i$-th layer and $\mathbf{W}_j^i$ is the $j$-th filter in the $i$-th layer. For the fully-connected layer, its weight matrix is indeed an exception of $h^i = 1$ and $w^i = 1$. 
In order to simplify the explanations, in the following contents, we only talk about the convolutional layers as illustrations.

Network pruning can be implemented by imposing a mask $\mathbf{T}^i$ upon $\mathbf{W}^i$. Here $\mathbf{T}^i$ is a binary tensor (0 or 1) with its entries indicating the states of network connections, \emph{i.e.}, whether the corresponding weights are pruned or not. Thus, given an expected pruning rate $p$, network pruning is formally expressed as:
\begin{equation}\label{network_sparsity}
\begin{split}
    &\mathop{\arg\max}_{\mathbf{T}^i}  \mathcal{L}(\mathbf{W}^i \oplus \mathbf{T}^i), \;\; s.t.  \;\; \frac{\| \mathbf{T}^i \|_0}{K} = 1 - p,
\end{split}
\end{equation}
where $\oplus$ represents the masking operation, $\mathcal{L}(\cdot)$ measures the importance of its input, and $K$ denotes the size of $\mathbf{T}^i$ that varies according to the basic pruning granularity. We measure the input importance using the $\ell_1$ norm of the basic pruning granularity. We find this criterion sufficient, however, other metrics, such as weight gradients~\cite{molchanov2016pruning}, activation sparsity~\cite{molchanov2017variational}, can be adopted as well.

\textbf{Weight Pruning}. The studies on weight pruning remove individual weights at any location of $\mathbf{W}^i$. Therefore, each mask $\mathbf{T}^i$ in weight pruning has the same shape with $\mathbf{W}^i$ of $\mathbb{R}^{n^i \times m^i \times h^i \times w^i}$ and its size $K = n^i \cdot m^i \cdot h^i \cdot w^i$.  The specific objective of weight pruning is:
\begin{equation}\label{weight_sparsity}
\begin{split}
    \mathop{\arg\max}_{\mathbf{T}^i} \sum_{j}^{n^i}  \sum_{k}^{m^i}&\sum_{q}^{h^i}\sum_{r}^{w^i} \mathcal{L}(\mathbf{W}^i_{j,k,q,r} \cdot \mathbf{T}^i_{j,k,q,r}), \\& s.t. \;\; \frac{\| \mathbf{T}^i \|_0}{K} = 1 - p.
\end{split}
\end{equation}

\textbf{Filter Pruning}. The studies on filter pruning remove the entire filter $\mathbf{W}^i_j$. Thus, each mask $\mathbf{T}^i$ in filter pruning has the shape of $\mathbb{R}^{n^i}$ and $K = n^i$. The specific objective of filter pruning is:
\begin{equation}\label{filter_sparsity}
\begin{split}
    &\mathop{\arg\max}_{\mathbf{T}^i}  \sum_{j=1}^{n^i} \mathcal{L}(\mathbf{W}^i_j \cdot \mathbf{T}^i_j), \;\; s.t. \;\; \frac{\| \mathbf{T}^i \|_0}{K} = 1 - p.
\end{split}
\end{equation}

In the following, we introduce a novel pattern of 1$\times$N pruning, whose basic pruning granularity falls into consecutive N output kernels with the same input channel index. We show that our 1$\times$N pruning can be an efficient and effective alternative to simultaneously accelerate the model inference on modern CPUs-based platforms and retain the accuracy performance.

\subsection{1$\times$N Pruning Pattern}\label{1xnblocksparsity}

We define the problem of pruning CNNs using our 1$\times$N pattern. In order to simplify the explanations, we reformat the representation of $\mathbf{W}^i \in \mathbb{R}^{n^i \times m^i \times h^i \times w^i}$ as $\mathbf{\Omega}^i \in \mathbb{R}^{m^i \times n^i}$. Note that each element in $\mathbf{\Omega}^i$ is a kernel with the shape of $h^i \times w^i$, \emph{i.e.}, $\mathbf{\Omega}^i_{k, j} = \mathbf{W}^i_{j, k, :, :}$. Each column $\mathbf{\Omega}^i_{:, j}$ stands for a filter and each row $\mathbf{\Omega}^i_{k, :}$ consists of these kernels that have the same input channel index of $k$. 

As shown in Fig.\,\ref{existing}, we further partition the whole $\mathbf{\Omega}^i$ into a collection of smaller blocks. Our partition can be made more precise for an $m^i \times n^i$ matrix $\mathbf{\Omega}^i$ by partitioning $m^i$ into a collection of $m^i$ row-groups, and then further partitioning $n^i$ into a collection of $\frac{n^i}{N}$ col-groups. Consequently, each block $(k, j)$ is a 1$\times$N matrix including consecutive N output kernels with the same input channel index $k$, namely ${\mathbf{\Omega}}^i_{k, (j-1) \cdot N + 1 \, : \, j \cdot N}$. Based on this partitioned matrix, the basic pruning granularity of our 1$\times$N sparsity falls into these blocks. Thus, the mask, $\mathbf{T}^i$, in our 1$\times$N pruning has the shape of $\mathbb{R}^{m^i \times \frac{n^i}{N}}$ and its size $K = m^i \cdot \frac{n^i}{N}$. Finally, the specific objective of our 1$\times$N pruning is:
\begin{equation}\label{block_sparsity}
\begin{split}
    \mathop{\arg\max}_{\mathbf{T}^i}  \sum_{k=1}^{m^i}&\sum_{j=1}^{\frac{n^i}{N}}\mathcal{L}({\mathbf{\Omega}}^i_{k, (j-1) \cdot N + 1 \, : \, j \cdot N} \cdot \mathbf{T}^i_{k, j}), \\& s.t. \;\; \frac{\| \mathbf{T}^i \|_0}{K} = 1 - p.
\end{split}
\end{equation}

Furthermore, we realize that weight pruning and filter pruning are two special cases of our proposed 1$\times$N pruning pattern. Specifically, our 1$\times$N pruning degenerates to weight pruning subject to N = 1, $h^i = 1$ and $w^i = 1$. Besides, it further degenerates to filter pruning if $N = n^i$. When $1 < N < n^i$, our method provides an intermediate granular level for network pruning, since it is coarser as compared to the fine-grained weight pruning but finer as compared to the coarse-grained filter pruning. Many previous researches~\cite{mao2017exploring,wortsman2019discovering,xie2019exploring} also explore removing kernels; however, our pruning pattern has a stronger requirement of continuity on the N removed kernels, with its merits in acceleration since these consecutive kernels can be stored continuously in the memory cache and the convolution with the inputs can proceed using a parallelized block-wise vectorized operation as analyzed in Sec.\,\ref{decoding_efficiency}. Therefore, it is expected that our 1$\times$N pruning can offer a better performance than the filter pruning, and also an apparent inference speedup compared to the weight pruning.

\begin{figure*}[!t]
\begin{center}
\includegraphics[height=0.40\linewidth]{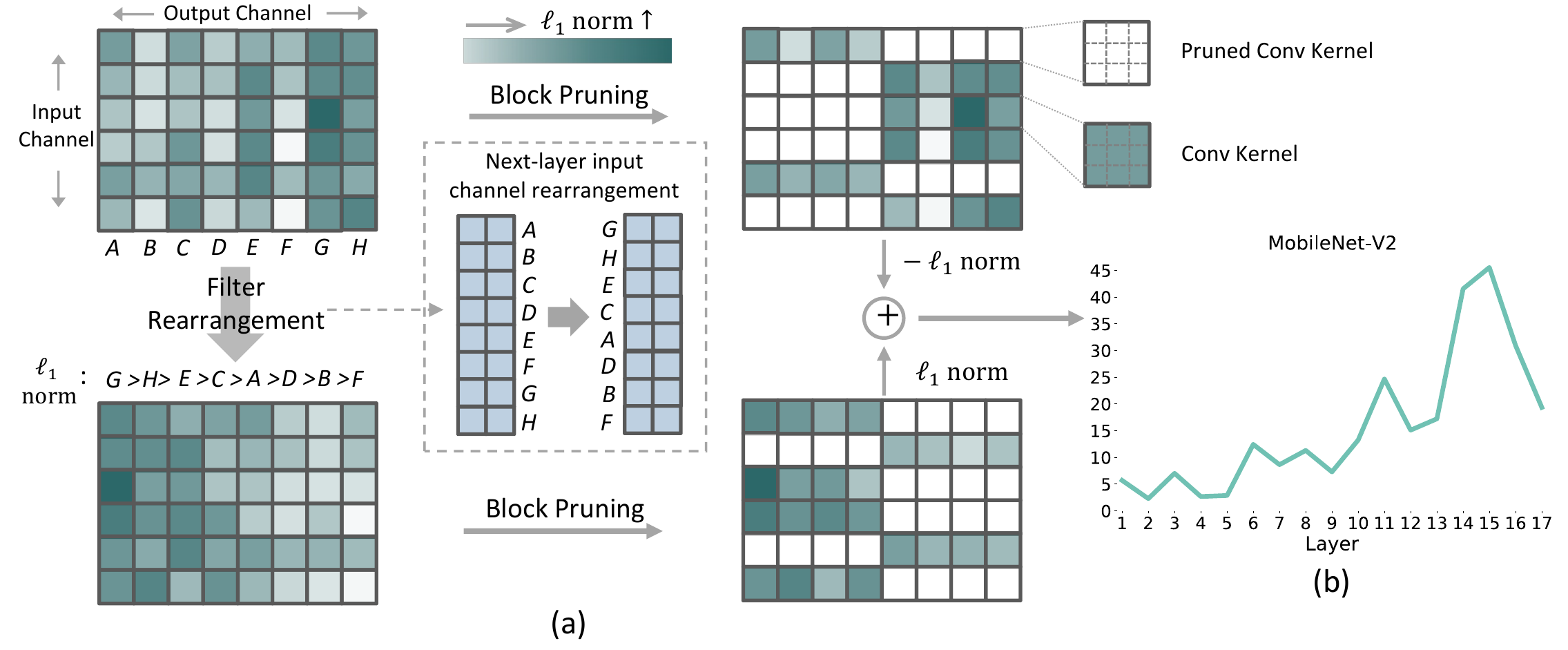}
\end{center}
\caption{\label{rearrange}
Workflow of filter rearrangement. We rearrange the weight matrix in the output channel dimension using the $\ell_1$ norm of each filter. Then, similar rearrangement is applied to the next-layer weight matrix in the input channel dimension. As results, more influential kernels with larger $\ell_1$ norms are preserved for accuracy improvements, as validated using MobileNet-V2. Best viewed in colors.
}
\end{figure*}

As a distinguished difference from weight pruning and filter pruning, pruning kernels provides an intermediate granular level for network sparsity, since it is coarser as compared to the fine-grained weight pruning but finer as compared to the coarse-grained filter pruning.

\subsection{Filter Rearrangement}\label{filter_rearrangement}

Our 1$\times$N pruning in this paper follows the typical three-step pipeline of network training, then pruning consecutive kernels with smaller $\ell_1$ norms, and fine-tuning the sparse one to recover the performance in the end. 
At the second pruning step, inspired by~\cite{mishra2021accelerating} which conducts channel permutations to preserve more high-magnitude weights in the N:M weight pruning, we realize that the layout of $\mathbf{\Omega}^i$ can be altered to further relieve the pruning impact.

To implement the above process, we propose a workflow of filter rearrangement, whose working principle is shown in Fig.\,\ref{rearrange}. Given the original weight matrix $\mathbf{\Omega}^i$ in the top-left of (a), simply applying 1$\times$N pruning upon $\mathbf{\Omega}^i$ leads to the loss of some kernels with relatively large values of $\ell_1$ norms in the top-right of (a). We calculate the $\ell_1$ norm of each filter, \emph{i.e.}, one column in $\mathbf{\Omega}^i$, then rearrange $\mathbf{\Omega}^i$ in the output channel dimension according to the calculated filter norm in a decreasing order as shown in the lower-left of (a). The rearranged weight matrix is denoted as $\tilde{\mathbf{\Omega}}^i$ to which our 1$\times$N pruning is further applied. As a consequence, more kernels with larger $\ell_1$ norms are preserved after pruning in the lower-right of (a), leading to an overall increasing weight magnitude as verified on MobileNet-V2 of (b). The filter rearrangement requires the outputs to be similarly rearranged as well, so as to maintain the same convolutional results with the next-layer weight matrix. However, frequently rearranging outputs incurs more run-time cost in the inference. Alternatively, we choose to apply a similar rearrangement to the input channel dimension of the next-layer weight matrix as illustrated in the middle of (a), which is accomplished once for all before pruning and thus brings no run-time cost.  The effectiveness of filter rearrangement for accuracy improvements is presented in Sec.\,\ref{experiments}.

Herein, we want to stress the difference of our filter rearrangement against the channel permutation~\cite{mishra2021accelerating}. First, rearranging the columns of $\mathbf{\Omega}^i$ is indeed to change the position of each filter in our setting. Second, our goal is to preserve more high-magnitude kernels, while~\cite{mishra2021accelerating} is to preserve more individual high-magnitude weights. Third, we {simply} accomplish our rearrangement according to the $\ell_1$ norm of each filter. {The implementation details of channel permutations are discussed in another paper~\cite{pool2021channel} where a bounded regressions-based permutation search is proposed to find a high-quality permutation.
}

Then, Eq.\,(\ref{block_sparsity}) after filter rearrangement is rewritten as:
\begin{equation}\label{new_block_sparsity}
\begin{split}
    \mathop{\arg\max}_{\mathbf{T}^i}  \sum_{k=1}^{m^i}&\sum_{j=1}^{\frac{n^i}{N}}\mathcal{L}(\tilde{{\mathbf{\Omega}}}^i_{k, (j-1) \cdot N + 1 \, : \, j \cdot N} \cdot \mathbf{T}^i_{k, j}), \;\; \\&
    s.t. \;\; \frac{\| \mathbf{T}^i \|_0}{K} = 1 - p.
\end{split}
\end{equation}

Note that the maximization of the above objective can be achieved by setting to $1$s the entries of $\mathbf{T}^i$ corresponding to these blocks in $\tilde{\mathbf{\Omega}}^i$ with their $\ell_1$ norms within the largest top-$(1 - p)$, and $0$s otherwise. As a consequence, the pruned weights after applying our 1$\times$N pruning pattern is then derived as:
\begin{equation}
    \bar{\mathbf{\Omega}}^i = \tilde{\mathbf{\Omega}}^i \oplus \mathbf{T}^i.
\end{equation}

\begin{figure*}[!t]
\begin{center}
\includegraphics[height=0.55\linewidth]{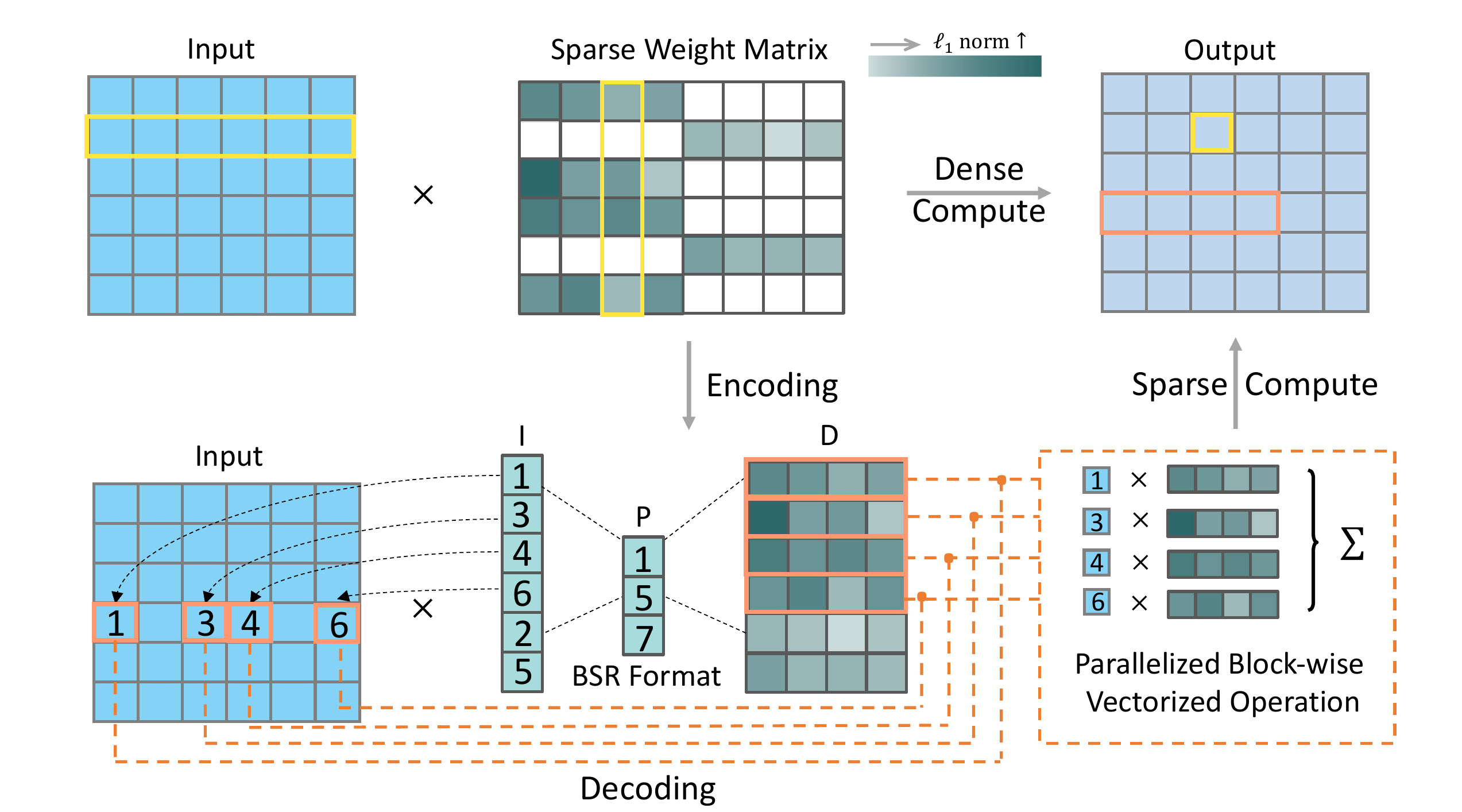}
\end{center}
\caption{\label{acceleratemechanism}
Encoding and decoding of our 1$\times$N pruning pattern. Our pruning pattern results in a sparse matrix with constant-size blocks, enabling it to be encoded by Block Compressed Sparse Row Format (BSR) to save the memory storage. In the decoding process, we calculate the outputs in a block-wise manner where the block-wise vectorized operation is applied in parallel to realize practical speedups. Best viewed in colors.
}
\end{figure*}

\subsection{Encoding and Decoding Efficiency}\label{decoding_efficiency}

Given the input activation tensor $\mathbf{X}^i$ as illustrated in Fig.\,\ref{acceleratemechanism}, the output tensor $\mathbf{Y}^i$ calculated by the standard dense-matrix structure is obtained as:
\begin{equation}\label{dense}
    \mathbf{Y}^i_{k, j} = \sum_{z = 1}^{m^i}\mathbf{X}^i_{k, z} \bar{\mathbf{\Omega}}^i_{z, j}.
\end{equation}

Considering a sparse matrix, operations using the standard dense-matrix structure bear inefficiency as processing and memory are wasted on a large number of zero-valued elements. Thus, it is of great necessity to take advantage of specialized data structures in order to store and manipulate the sparse matrix. However, the irregular sparse matrix resulting from weight pruning requires a great many of indices to record the positions of the reserved weights. Though Compressed Sparse Row Format (CSR) can be used to save index storage, irregular sparsity barely takes advantage of vector processing architectures and memory buses, resulting in little acceleration and even speed deterioration.

Weight pruning leads to an irregular sparse matrix, therefore, a large number of indices are needed to record the positions of the reserved weights. To save the index storage, the relative Compressed Sparse Row (CSR) format is usually adopted~\cite{han2015learning,han2016deep}, which encodes each index by the relative distance (\emph{i.e.}, the number of zeros) between two adjacent non-zero weights. Besides, a decoding process is needed to select the corresponding activations for the reserved weights. The main drawback of irregular pruning is that decoding one index requires a search over the whole activation vector, thus it brings little acceleration, even speed degradation.

In contrast, due to the requirement of continuity on the N removed kernels, our pruning pattern results in a sparse matrix $\bar{\mathbf{\Omega}}^i$ with constant-sized blocks. This good property brings two merits: 
First, the constant-sized blocks are by nature more easily encoded by Block Compressed Sparse Row Format (BSR)~\cite{shahnaz2011blocked} to save non-zero elements in $\bar{\mathbf{\Omega}}^i$ with significantly less storage for the indices. 
Second, the output tensor $\mathbf{Y}^i$ is derived using a block-wise vectorized operation in parallel to achieve an apparent speedup.
Specifically, as illustrated in Fig.\,\ref{acceleratemechanism}, the pruned weight matrix $\bar{\mathbf{\Omega}}^i$ in our 1$\times$N pruning is encoded by BSR into three components: $1) \mathbf{D}^i \in \mathbb{R}^{t \times N}$, $2) \mathbf{I}^i \in \mathbb{R}^t$ and $3) \mathbf{P}^i \in \mathbb{R}^{\frac{n^i}{N} + 1}$, where $t$ is the number of non-zero blocks in $\bar{\mathbf{\Omega}}^i$. The matrix, $\mathbf{D}^i$, and vector, $\mathbf{I}^i$, contain non-zero blocks and their row indices in $\bar{\mathbf{\Omega}}^i$, respectively. We form $\mathbf{D}^i$ by firstly stacking up non-zero blocks within the same col-group of $\bar{\mathbf{\Omega}}^i$, and concatenating the stacked ones across different col-groups. The vector $\mathbf{I}^i$ records the row index of each block item in $\bar{\mathbf{\Omega}}^i$. The $k$-th element of the vector $\mathbf{P}^i$ encodes the start row index of the $k$-th col-group in $\mathbf{D}^i$. The last element of $\mathbf{P}^i$ is a fictitious index, which is always equal to $t + 1$. Attributed to the block storage format of $\mathbf{D}^i$, we can calculate  $\mathbf{Y}^i$ in a block-wise manner during decoding as follows:
\begin{equation}\label{vectorized}
    \mathbf{Y}^i_{k,\, (j-1) \cdot N + 1 : j \cdot N} = \sum_{z = \mathbf{P}^i_{ j/N }}^{\mathbf{P}^i_{ j/N + 1} - 1} \mathbf{X}^i_{k, \mathbf{I}^i_z} \cdot \mathbf{D}^i_{\mathbf{I}^i_z,:}.
\end{equation}

\begin{table*}[!th]
    \caption{Performance studies of our 1$\times$N pruning with and without filter rearrangement. The experiment is conducted using MobileNet-V2 with the pruning rate $p=50\%$.}
    \centering
    \begin{tabular}{c|c|c|c|c|c|c|c|c|c|c}
    \hline
    & \multicolumn{2}{c|}{N=2 (\%)}     & \multicolumn{2}{c|}{N=4 (\%)} & \multicolumn{2}{c|}{N=8 (\%)} & \multicolumn{2}{c|}{N=16 (\%)} & \multicolumn{2}{c}{N=32 (\%)} \\\hline
    & Top-1 & Top-5 & Top-1 & Top-5 & Top-1 & Top-5 & Top-1 & Top-5  & Top-1 & Top-5         \\\hline
w/o Rearrange    & 69.900 & 89.296                     & 69.521         & 88.920         & 69.206          & 88.608        & 68.971          & 88.399         & 68.431          & 88.315         \\\hline
Rearrange & 70.233 & 89.417 & 69.579         & 88.944        & 69.372        & 88.862        & 69.352          & 88.708         & 68.762 & 88.425 \\\hline
    \end{tabular}
    \label{w/orearrange}
\end{table*}
\begin{table}[!t]
    \caption{{Performance studies of our 1$\times$N pruning with kernel-wise pruning. The experiment is conducted using MobileNet-V2 and ResNet-50 with the pruning rate $p=50\%$.}}
    \centering
    \begin{tabular}{c|c|c|c|c}
    \hline
    & \multicolumn{2}{c|}{{ResNet-50 (\%)}}     & \multicolumn{2}{c}{{MobileNet-V2 (\%)}}  \\\hline
    & {Top-1} &{Top-5} &{Top-1} &{Top-5}        \\\hline
  {1$\times$N (N=4)}  &{76.506} &{93.238} & {69.706}         &{ 89.165}      \\\hline
{kernel (random)}    & {74.834} & {92.178}                    & {68.615}         & {88.434}        \\\hline
{kernel ($\ell_1$)}  & {75.370} & {92.582} & {69.514}         & {89.012}            \\\hline
    \end{tabular}
    \label{kernel-comparison}
\end{table}

With proper index vectors $\mathbf{P}^i$ and $\mathbf{I}^i$, we can directly fetch these activations corresponding to non-zero weights for the output computation, through which we avoid a complete involvement of the whole activation tensor as with the dense-matrix structure. Besides, the block storage format of $\mathbf{D}^i$ also allows fast row data access since each block $\mathbf{D}^i_{\mathbf{I}^i_z,:}$ is stored continuously in the memory. Besides, the block-wise vectorized operation in Eq.\,(\ref{vectorized}) can be implemented extremely fast as the multiplication between input item $\mathbf{X}^i_{k, \mathbf{I}^i_z}$ and each entry of $\mathbf{D}^i_{\mathbf{I}^i_z,:}$ can proceed in parallel. Thus, our 1$\times$N pruning enables apparent acceleration on the general CPUs-based devices. To ensure end-to-end execution efficiency, we utilize the optimizing compiler TVM~\cite{chen2018tvm} to enable optimal code generation. And based on Eq.\,(\ref{vectorized}), we use Ansor~\cite{zheng2020ansor} for automated tensor program generation to search best sparse convolution implementation.

\section{Experiments}\label{experiments}

\subsection{Implementation Settings}\label{implementation}

For fair comparison, similar to our 1$\times$N pruning, we {re-implement} the compared baselines of weight pruning and filter pruning using $\ell_1$ norm as the importance evaluation. Besides, given the expected pruning rate $p$, we simply remove per-layer weights/filters/blocks with their corresponding $\ell_1$ norms within the smallest top-$p$. All experiments are performed using the pre-trained light-weight MobileNet-V1~\cite{howard2017mobilenets}, -V2~\cite{sandler2018mobilenetv2}, -V3~\cite{howard2019searching} and large-scale ResNet-50~\cite{he2016deep} on ILSVRC-2012~\cite{deng2009imagenet} that contains over $1.2$ million images for training and $50,000$ validation images from $1,000$ classes. After pruning, we fine-tune the sparse models for 90 epochs on two NVIDIA V100 GPUs with the settings: Stochastic Gradient Descent (SGD) optimizer with a momentum of 0.9, weight decay of 4e-5 for MobileNets and 1e-4 for ResNet-50, and initial learning rate of 0.1 with a cosine annealing. The data augmentation includes random cropping and horizontal flipping.

\subsection{Ablation Study}\label{ablation}
We first analyze the influence of filter rearrangement in Sec.\,\ref{filter_rearrangement}. Table\,\ref{w/orearrange} compares the performance of our 1$\times$N pruning for pruning MobileNet-V2 with and without filter rearrangement. The pruning rate $p$ is set to $50\%$. As can be seen, rearranging filters consistently enhances the accuracy performance in both the top-1 and top-5, even with various block size (N). For example, the top-1 classification accuracy of pruned MobileNet-V2 with 1$\times$16 pruning is increased by $0.381\%$ (69.352\% with and 68.971\% without filter rearrangement). To dive into a deeper analysis, by rearranging the weight matrix in the output channel dimension, more blocks with larger $\ell_1$ norms are preserved after applying our 1$\times$N as validated in Fig.\,\ref{rearrange}(b). These results well validate the effectiveness of filter rearrangement in boosting the performance of pruned models.
{We continue the study on our consecutive kernel removal. Recall that consecutive N output kernels with the same input channel index are grouped into one block and our 1$\times$N removes these blocks with a smaller $\ell_1$ norm. In Table\,\ref{kernel-comparison}, we compare our consecutive kernel removal with two variants including (1) removing kernels randomly and (2) removing kernels with smaller $\ell_1$ magnitudes. From these results in Table\,\ref{kernel-comparison}, we can see that removing random kernels performs worse than removing smaller magnitude kernels, indicating the importance of preserving larger magnitude weights.
Our stronger constraint of block-level magnitude leads to performance increase compared with the kernel-level magnitude. Also, our 1$\times$N merits in its acceleration since the consecutive kernels can be stored continuously in the memory cache. Moreover, the convolution with the inputs can proceed using a parallelized block-wise vectorized operation as analyzed in Sec.\,\ref{decoding_efficiency} and verified in the following Sec.\,\ref{comparison}.

}

\begin{table*}[!t]
    \caption{Performance comparison of our 1$\times$N pruning against weight pruning and filter pruning. The experiment is conducted using MobileNet-V1/-V2/-V3 {and ResNet-50 with} the pruning rate $p=50\%$.}
    \centering
    \begin{tabular}{c|c|c|c|c|c|c|c|c|c|c}
    \hline
     & \multicolumn{8}{c|}{MobileNet (p = 50\%)} & \multicolumn{2}{c}{{ResNet-50 (p = 50\%)}} \\ \hline
    \multirow{2}{*}{} & \multicolumn{2}{c|}{V1 (\%)} & \multicolumn{2}{c|}{V2 (\%)} & \multicolumn{2}{c|}{V3-small (\%)} &\multicolumn{2}{c|}{V3-large (\%)} &\multicolumn{2}{c}{{(\%})}\\ \cline{2-11} 
     & Top-1 & Top-5  & Top-1 & Top-5 & Top-1 & Top-5  & Top-1 & Top-5 & {Top-1} & {Top-5}\\ \hline\hline
     Origin &71.154  &89.834  &71.737 &90.452 &67.225  &87.351 &74.280 &91.928 &{77.008}  &{93.654}\\ \hline\hline
    Weight Pruning &70.764  &89.592   & 71.146 & 89.872 &66.376 &86.868  &72.897 &91.093 &{77.088}  &{93.614}\\ \hline\hline
    Filter Pruning &65.348  &86.264   & 66.730 & 87.190 &59.054  &81.743  &69.137 &89.097 &{75.382}  &{92.518}\\ \hline\hline
    1$\times$2 Pattern (\textbf{Ours}) &70.281 &89.370  &70.233 &89.417  &65.380  &86.060 &72.120 &90.677 &{76.654} &{93.466}\\ \hline
    1$\times$4 Pattern (\textbf{Ours}) &70.052 &89.056  &69.706  &89.165  &64.465  &85.495 &71.935 &90.458 &{76.506} &{93.238} \\ \hline
    1$\times$8 Pattern (\textbf{Ours}) &69.908 &89.027  &69.372  &88.862  &64.101  &85.274 &71.478 &90.163 &{76.146} &{93.134}\\ \hline
    1$\times$16 Pattern (\textbf{Ours}) &69.559 &88.933  &69.352  &88.708  &63.126  &84.203  &71.112 &90.129 &{76.254} &{93.084}\\ \hline
    1$\times$32 Pattern (\textbf{Ours}) &69.541 &88.801  &68.762  &88.425  &62.881  &83.982 & 70.769&89.696 &{75.960} &{92.950}\\ 
    \hline
    \end{tabular}
    \label{p50}
\end{table*}

\subsection{Performance Comparison}\label{comparison}

In this section, we compare the performance of our proposed 1$\times$N pruning with traditional weight pruning and filter pruning. We show that the advantages of pruning pattern are reflected from two perspectives: $1$) maintaining better accuracy than filter pruning, and $2$) achieving apparent CPUs acceleration compared to weight pruning.

\begin{figure*}[!t]
\centering
\begin{minipage}[t]{0.45\textwidth}
\centering
\includegraphics[width=2.5in]{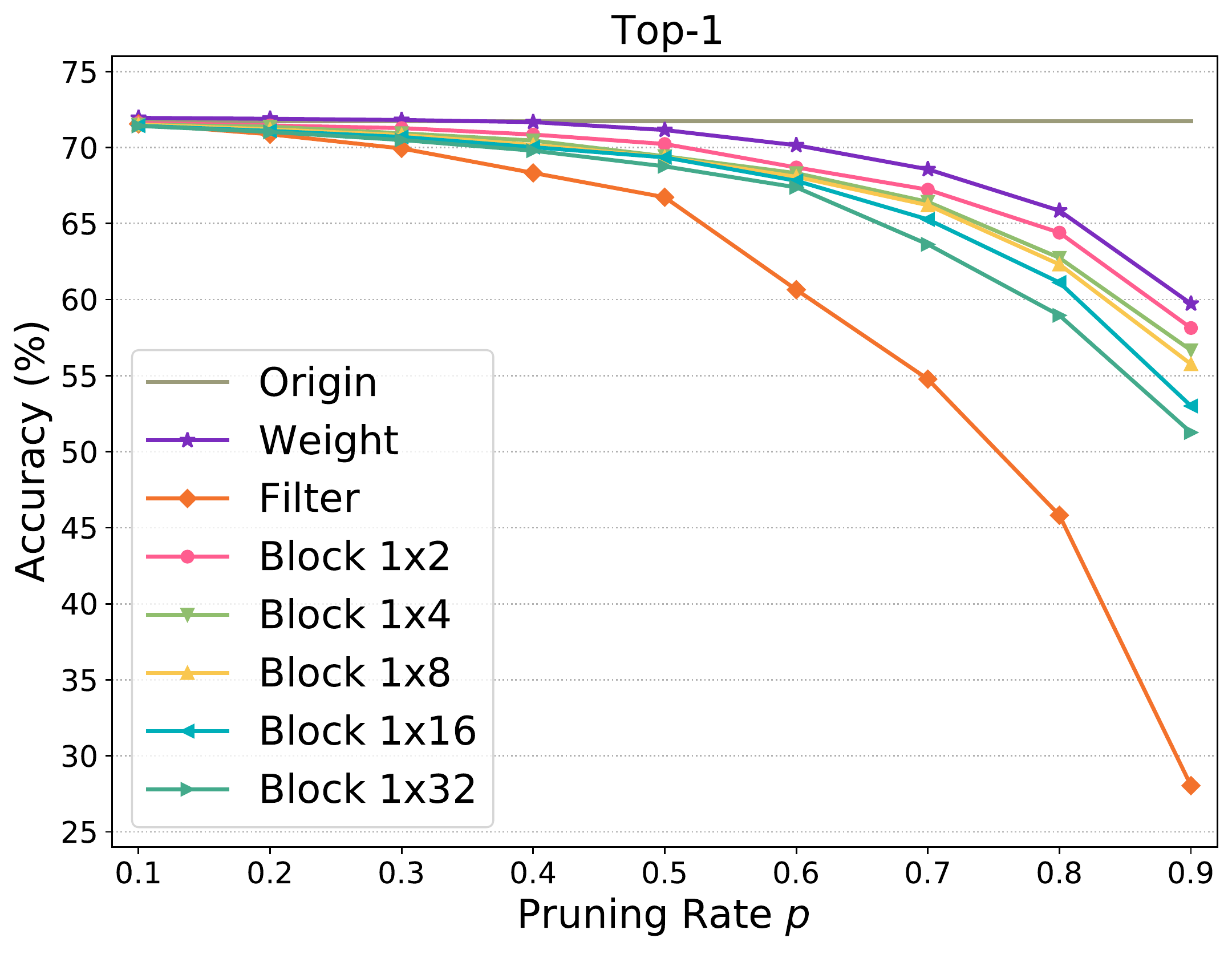}
\includegraphics[width=2.5in]{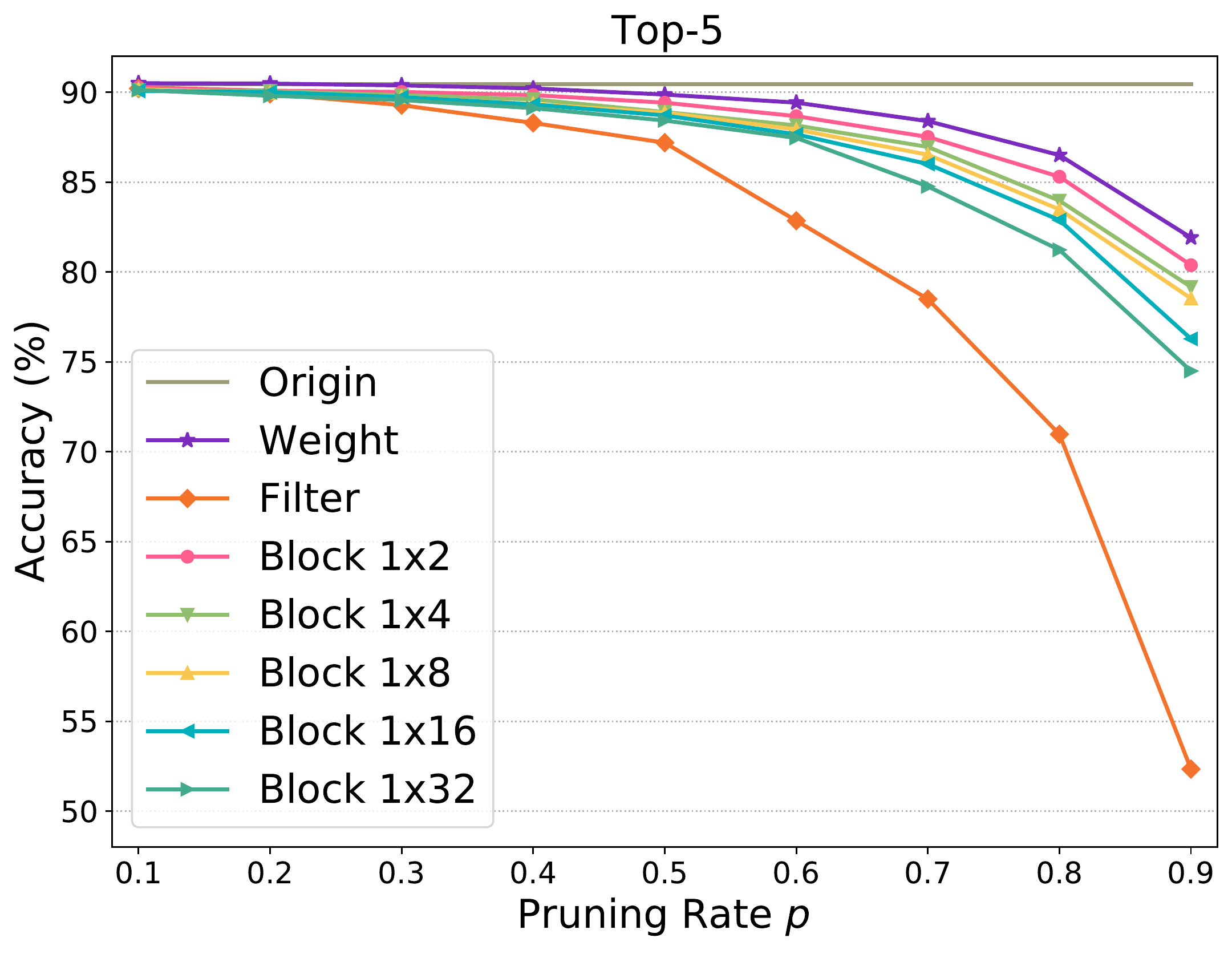}
\centerline{(a) MobileNet-V2}
\end{minipage}
\begin{minipage}[t]{0.45\textwidth}
\centering
\includegraphics[width=2.5in]{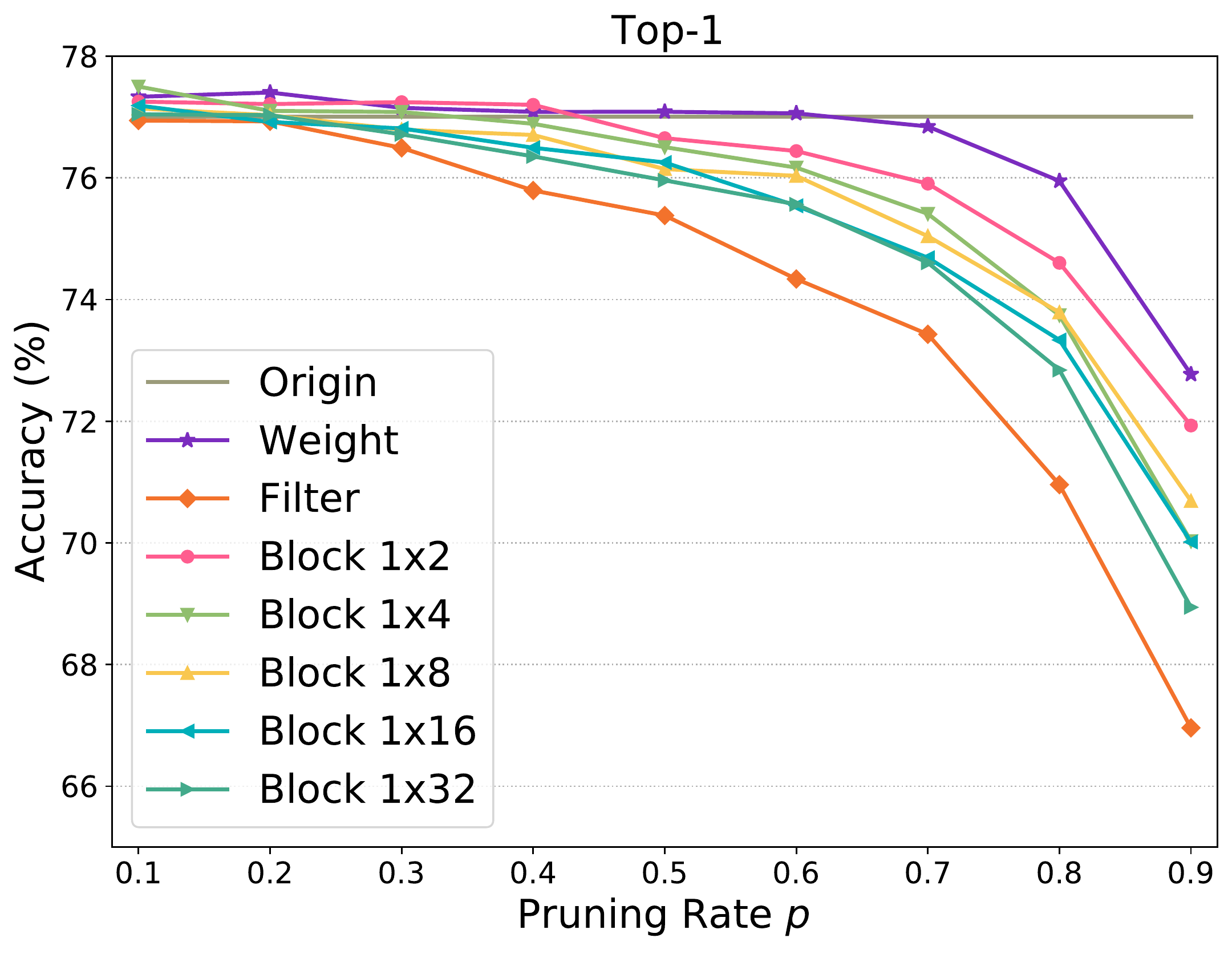}
\includegraphics[width=2.5in]{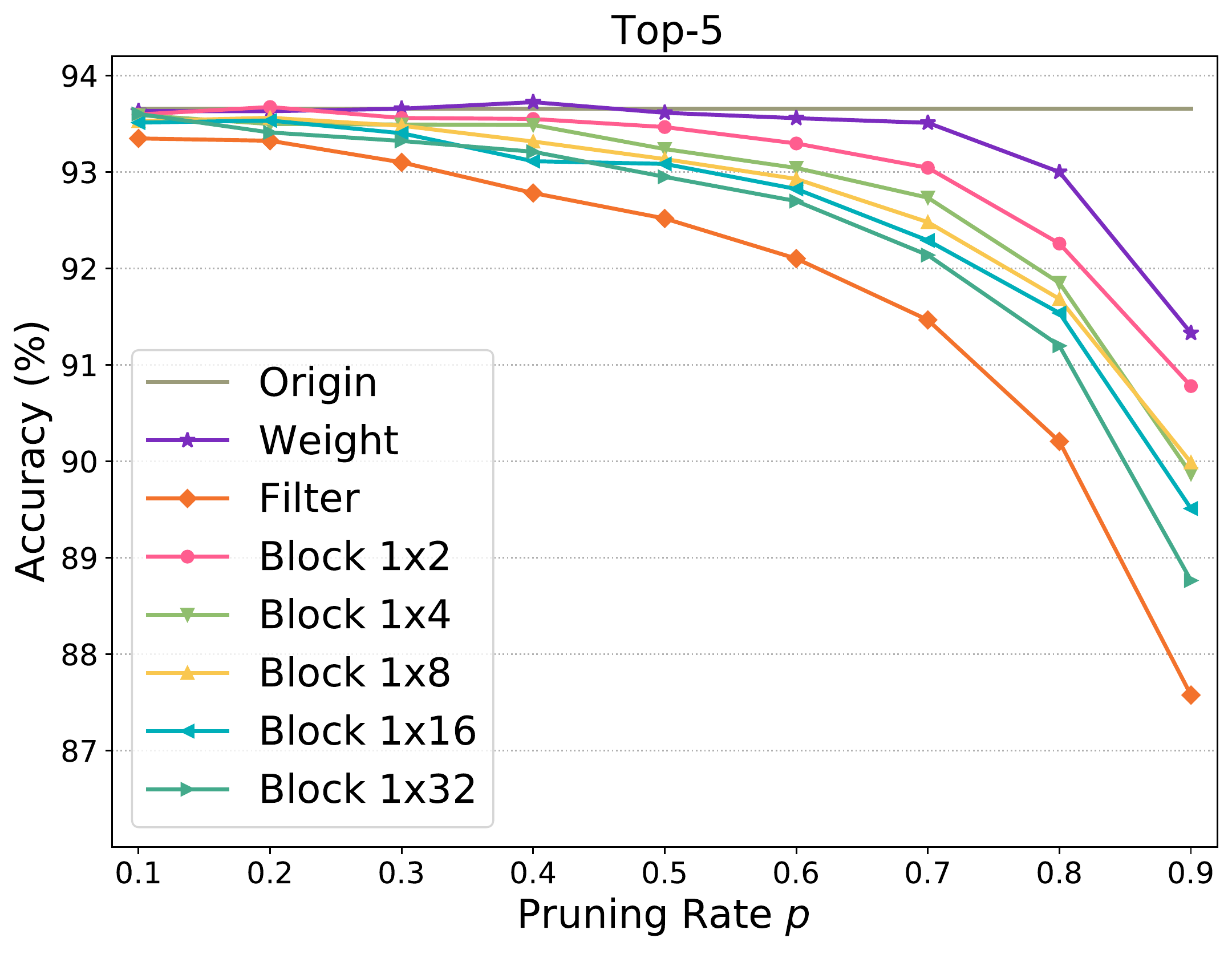}
\centerline{(b) ResNet-50}
\end{minipage}
\caption{\label{various_p}Performance comparison of our 1$\times$N pruning against weight pruning and filter pruning under different pruning rates. The experiment is conducted using MobileNet-V2 (left) and ResNet-50 (right). Best viewed in colors.}
\end{figure*}

\textbf{Accuracy Performance}. We first study the performance of 1$\times$N sparsity across different networks. Table\,\ref{p50} displays the pruning results of our 1$\times$N pruning and existing weight pruning and filter pruning using MobileNet-V1/-V2/-V3 with the pruning rate $p$ set to $50\%$. Table\,\ref{p50} shows that filter pruning suffers the most performance degradation of $5.806\%$, $5.007\%$, $8.171\%$, and $5.143\%$ when pruning MobileNet-V1, V2, V3-small and V3-large, respectively. Such severe performance losses are attributed to its coarse-grained pruning granularity. Consequently, the poor performance barricades the using of filter pruning in practical model deployment. In contrast, due to its fine-grained pruning granularity, weight pruning presents the best performance with top-1 accuracy losses of $0.390\%$, $0.591\%$, $0.849\%$, and $1.383\%$ when pruning MobileNet-V1, V2, V3-small and V3-large, respectively. Despite its ability to maintain high accuracy, weight pruning achieves rare acceleration as analyzed in the following. The poor speedup also disables the using of  filter pruning. With respect to our proposed 1$\times$N pruning, we have two observations: $1$) our method well boosts the performance of filter pruning regardless of the block size N. Taking MobileNet-V2 as an example, our 1$\times$4 pruning achieves $69.706\%$ top-1 accuracy, significantly better than $66.730\%$ of filter pruning. Though it is slightly poorer than $71.146\%$ of weight pruning, our 1$\times$4 pruning obtains an apparent speedup as detailed in the following context. $2$) The performance of 1$\times$N pruning degenerates as the block size N increases. The rationale behind this is that {a} larger N indicates coarser pruning. As analyzed in Sec.\,\ref{1xnblocksparsity}, our 1$\times$N pruning degenerates to weight pruning with a small N and filter pruning with a large N.

{Table\,\ref{p50} also provides the performance comparison when using ResNet-50 as the network backbone with the pruning rate $p=50\%$. We can observe similar phenomena to MobileNets that filter pruning suffers the most performance drops and weight pruning presents best performance while our 1$\times$N provides a trade-off. Besides, our performance decreases as the block size N increases.
}

Fig.~\ref{various_p} further shows the performance comparison when applying different pruning rates to sparsifying MobileNet-V2 and ResNet-50. We can see that the increasing pruning rate results in decreasing accuracy performance for all methods. However, filter pruning degrades drastically if $p > 50\%$. In contrary, our pruning pattern maintains a similar decreasing tendency and close performance to weight pruning even if the pruning rate $p$ is very high\footnote{{The raw data for plotting Fig.~\ref{various_p} can be found from our project at \url{https://github.com/lmbxmu/1xN}}.}.

\begin{figure*}[!t]
\centering
\begin{minipage}[t]{0.48\textwidth}
\centering
\includegraphics[width=2.5in]{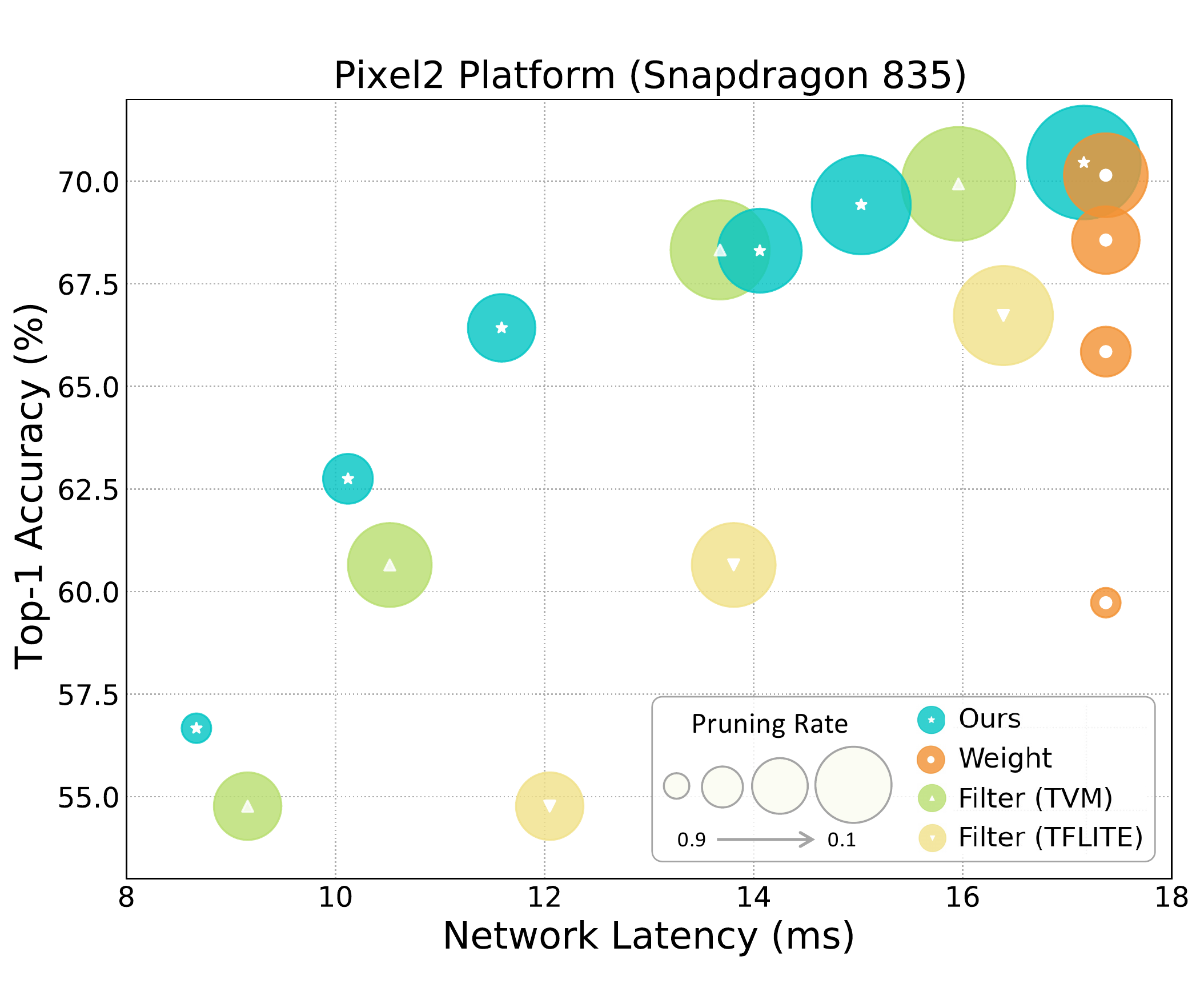}
\end{minipage}
\begin{minipage}[t]{0.48\textwidth}
\centering
\includegraphics[width=2.5in]{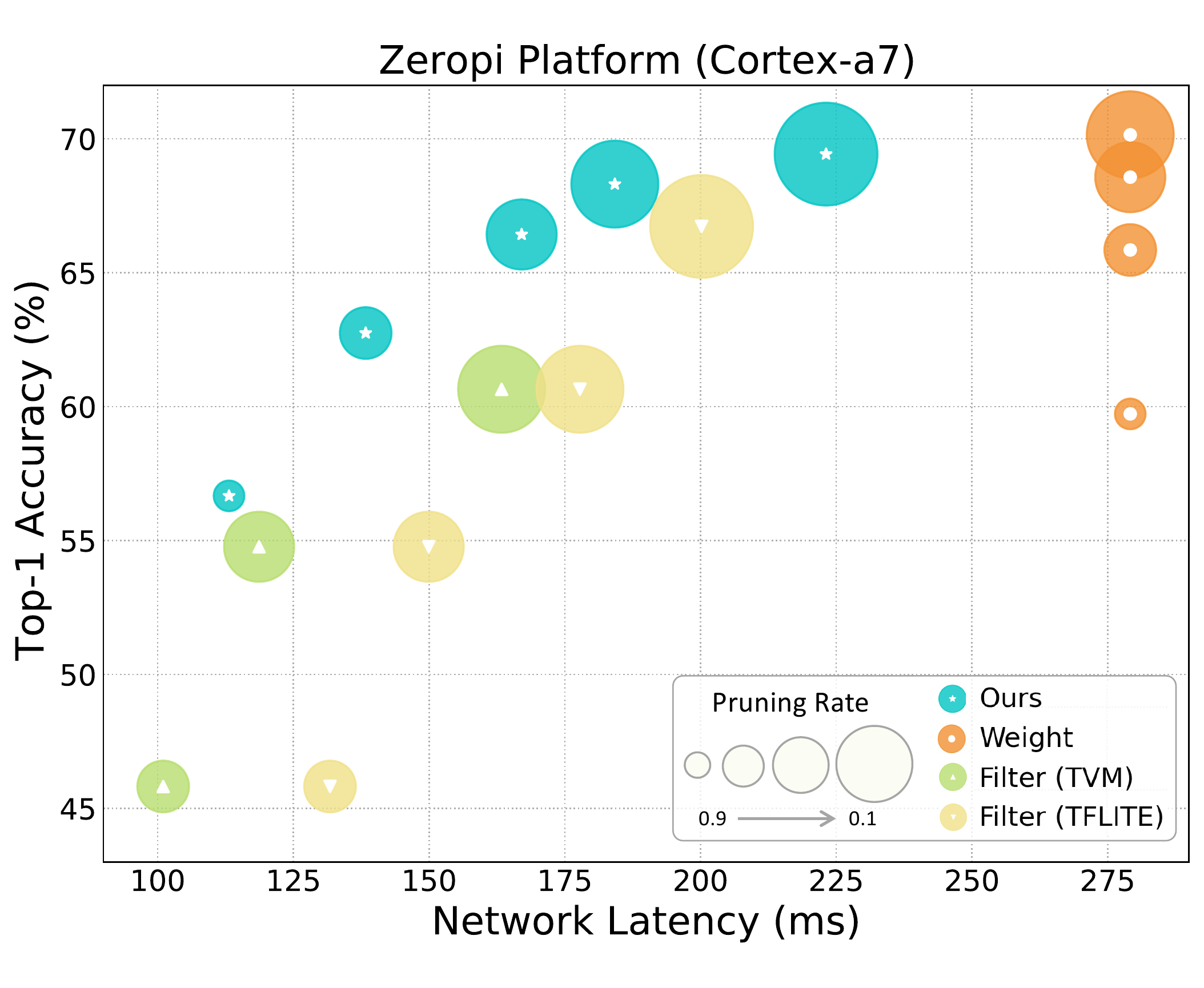}
\end{minipage}
\caption{\label{latency}Performance and latency comparison between our 1$\times$N (N=4) pruning against weight pruning and filter pruning. The experiment is conducted using MobileNet-V2 on the mobile platform of Pixel2 equipped with a Snapdragon 835 CPU (left), and the embedded platform of Zeropi equipped with a Cortex-a7 CPU (right). Best viewed in colors.}
\end{figure*}

\textbf{CPUs Acceleration}. 
Fig.\,\ref{latency} presents the experimental results, which are conducted to further explore the acceleration capacity of different methods on CPUs-based platforms. In Sec.\,\ref{decoding_efficiency}, we adopt TVM~\cite{chen2018tvm} to compile the pruned model of our pruning pattern. For fair comparison, we also consider TVM compiler for weight pruning and filter pruning. Besides, additional experiments by TFLite compiler~\cite{tflite2019} are also presented for weight pruning and filter pruning. After model compiling, we respectively deploy the sparse models to obtain network latencies on the mobile platform of Pixel2 equipped with a Snapdragon 835 CPU and the embedded platform of Zeropi equipped with a Cortex-a7 CPU.

From Fig.\,\ref{latency}, we observe no speedup from weight pruning, despite its ability to preserve good performance. As analyzed in Sec.\,\ref{introduction}, weight pruning leads to irregular sparsity that hardly utilizes the vector processing architectures and memory buses. Thus, weight pruning often results in little acceleration and even speed deterioration. Filter pruning leads to the most significant speedups since it does not modify the network structure, so that the pruned network can be well fitted by regular hardware to achieve acceleration. Nevertheless, the severe performance degradation disables the using of filter pruning in the model deployment. In contrast, our 1$\times$N (N=4) pruning achieves noticeable latency reductions across various pruning rates such as $56.04$ms inference savings on Cortex-A7 CPU over weight pruning when $p = 50\%$, while maintaining comparable top-1 accuracy performance. Compared with weight pruning and filter pruning, our 1$\times$N pruning shows a better capacity of keeping a trade-off between latency and performance.

\section{Limitations}\label{limitation}

First, our filter rearrangement using $\ell_1$ norms of filters does not always guarantee maximizing magnitude. Though it does not seem to have happened in the experiments, opportunity exists that lower-magnitude kernels are retained. The approach to cluster large values as in~\cite{ji2018tetris} has no this issue. However, we observe a similar performance of the cluster against our rearrangement. We adopt the $\ell_1$ norm based rearrangement since it is much easier to implement.

Second, though our 1$\times$N is originally proposed for CNNs, we observe rare speedups on recurrent neural networks (RNNs) compared to these RNNs acceleration~\cite{narang2017block,kalchbrenner2018efficient}. Nevertheless, we are making efforts to break this limitation and expecting that our method can be well generalized to a wider variety of networks in the near future.

Third, this paper misses comparisons to many studies that also explore an intermediate pruning granularity~\cite{elsen2020fast,mao2017exploring,vooturi2018hierarchical,ji2018tetris}, most of which only report the theoretical acceleration in their papers. In our deployment, we find most of them fail to obtain practical acceleration, or only reach few CPU speedups in  a pruning rate of over 90\%~\cite{elsen2020fast}. Currently, we are not sure if something is wrong in our implementation of these methods. Our future work will focus on this topic to show that our method is worth building on.

Fourth, we do not introduce a new pruning {criterion} but use the $\ell_1$ norm as our measure to reflect the importance of these consecutive kernels.
This is because we find that existing pruning criteria show similar performance if fair training settings are given, which is also discussed in~\cite{liu2018rethinking, le2021network}. Thus, we focus on designing a new pruning pattern. However, it is unclear if a specialized pruning criterion exists in our 1$\times$N pattern. We will continue excavating this issue.

\section{Conclusion}\label{conclusion}
We introduce a novel 1$\times$N pruning pattern to simultaneously maintain model accuracy and achieve significant speedups on general CPUs. Unlike previous approaches that prune the individual weights or the whole filters, we design a pruning pattern that supports network pruning by removing consecutive N output kernels with the same input channel index. To preserve more influential kernels, we propose a workflow of filter rearrangement that rearranges the weight matrix in the output channel dimension and applies similar rearrangement to the next-layer weight matrix in the input channel dimension. Our pruning pattern leads to a sparse matrix with constant-sized blocks enabling the computation outputs by using a parallelized block-wise vectorized operation. The experiments on the MobileNets/ResNet-50 and CPUs embedded hardware platforms demonstrate the efficacy of our approach.

\section*{Acknowledgments}
This work was supported by the National Science Fund for Distinguished Young Scholars (No. 62025603), the National Natural Science Foundation of China (No. U21B2037, No. 62176222, No. 62176223, No. 62176226, No. 62072386, No. 62072387, No. 62072389, and No. 62002305), Guangdong Basic and Applied Basic Research Foundation (No. 2019B1515120049), and the Natural Science Foundation of Fujian Province of China (No. 2021J01002).


\ifCLASSOPTIONcaptionsoff
  \newpage
\fi




\bibliographystyle{IEEEtran}
\bibliography{main}

\begin{thebibliography}{10}
\providecommand{\url}[1]{#1}
\csname url@samestyle\endcsname
\providecommand{\newblock}{\relax}
\providecommand{\bibinfo}[2]{#2}
\providecommand{\BIBentrySTDinterwordspacing}{\spaceskip=0pt\relax}
\providecommand{\BIBentryALTinterwordstretchfactor}{4}
\providecommand{\BIBentryALTinterwordspacing}{\spaceskip=\fontdimen2\font plus
\BIBentryALTinterwordstretchfactor\fontdimen3\font minus
  \fontdimen4\font\relax}
\providecommand{\BIBforeignlanguage}[2]{{%
\expandafter\ifx\csname l@#1\endcsname\relax
\typeout{** WARNING: IEEEtran.bst: No hyphenation pattern has been}%
\typeout{** loaded for the language `#1'. Using the pattern for}%
\typeout{** the default language instead.}%
\else
\language=\csname l@#1\endcsname
\fi
#2}}
\providecommand{\BIBdecl}{\relax}
\BIBdecl

\bibitem{Sun2014DeepLF}
Y.~Sun, Y.~Chen, X.~Wang, and X.~Tang, ``Deep learning face representation by
  joint identification-verification,'' in \emph{Proceedings of the Advances in
  Neural Information Processing Systems (NeurIPS)}, 2014, pp. 1988--1996.

\bibitem{he2016deep}
K.~He, X.~Zhang, S.~Ren, and J.~Sun, ``Deep residual learning for image
  recognition,'' in \emph{Proceedings of the IEEE/CVF Conference on Computer
  Vision and Pattern Recognition (CVPR)}, 2016, pp. 770--778.

\bibitem{Kim2016CharacterAwareNL}
Y.~Kim, Y.~Jernite, D.~Sontag, and A.~M. Rush, ``Character-aware neural
  language models,'' in \emph{Proceedings of the AAAI Conference on Artificial
  Intelligence (AAAI)}, 2016, pp. 2741--2749.

\bibitem{han2015learning}
S.~Han, J.~Pool, J.~Tran, and W.~Dally, ``Learning both weights and connections
  for efficient neural network,'' in \emph{Proceedings of the Advances in
  Neural Information Processing Systems (NeurIPS)}, 2015, pp. 1135--1143.

\bibitem{evci2020rigging}
U.~Evci, T.~Gale, J.~Menick, P.~S. Castro, and E.~Elsen, ``Rigging the lottery:
  Making all tickets winners,'' in \emph{Proceedings of the International
  Conference on Machine Learning (ICML)}, 2020, pp. 2943--2952.

\bibitem{molchanov2016pruning}
P.~Molchanov, S.~Tyree, T.~Karras, T.~Aila, and J.~Kautz, ``Pruning
  convolutional neural networks for resource efficient inference,'' in
  \emph{Proceedings of the International Conference on Learning Representations
  (ICLR)}, 2017.

\bibitem{choquette2021nvidia}
J.~Choquette, W.~Gandhi, O.~Giroux, N.~Stam, and R.~Krashinsky, ``Nvidia a100
  tensor core gpu: Performance and innovation,'' \emph{IEEE Micro}, vol.~41,
  no.~02, pp. 29--35, 2021.

\bibitem{zhou2021learning}
A.~Zhou, Y.~Ma, J.~Zhu, J.~Liu, Z.~Zhang, K.~Yuan, W.~Sun, and H.~Li,
  ``Learning n:m fine-grained structured sparse neural networks from scratch,''
  in \emph{Proceedings of the International Conference on Learning
  Representation (ICLR)}, 2021.

\bibitem{hubara2021accelerated}
I.~Hubara, B.~Chmiel, M.~Island, R.~Banner, S.~Naor, and D.~Soudry,
  ``Accelerated sparse neural training: A provable and efficient method to find
  n: M transposable masks,'' \emph{arXiv preprint arXiv:2102.08124}, 2021.

\bibitem{mishra2021accelerating}
A.~Mishra, J.~A. Latorre, J.~Pool, D.~Stosic, D.~Stosic, G.~Venkatesh, C.~Yu,
  and P.~Micikevicius, ``Accelerating sparse deep neural networks,''
  \emph{arXiv preprint arXiv:2104.08378}, 2021.

\bibitem{renda2020comparing}
A.~Renda, J.~Frankle, and M.~Carbin, ``Comparing rewinding and fine-tuning in
  neural network pruning,'' in \emph{Proceedings of the International
  Conference on Learning Representation (ICLR)}, 2020.

\bibitem{li2016pruning}
H.~Li, A.~Kadav, I.~Durdanovic, H.~Samet, and H.~P. Graf, ``Pruning filters for
  efficient convnets,'' in \emph{Proceedings of the International Conference on
  Learning Representations (ICLR)}, 2016.

\bibitem{liu2019metapruning}
Z.~Liu, H.~Mu, X.~Zhang, Z.~Guo, X.~Yang, K.-T. Cheng, and J.~Sun,
  ``Metapruning: Meta learning for automatic neural network channel pruning,''
  in \emph{Proceedings of the IEEE/CVF International Conference on Computer
  Vision (ICCV)}, 2019, pp. 3296--3305.

\bibitem{guo2020dmcp}
S.~Guo, Y.~Wang, Q.~Li, and J.~Yan, ``Dmcp: Differentiable markov channel
  pruning for neural networks,'' in \emph{Proceedings of the IEEE/CVF
  Conference on Computer Vision and Pattern Recognition (CVPR)}, 2020, pp.
  1539--1547.

\bibitem{lin2020hrank}
M.~Lin, R.~Ji, Y.~Wang, Y.~Zhang, B.~Zhang, Y.~Tian, and L.~Shao, ``Hrank:
  Filter pruning using high-rank feature map,'' in \emph{Proceedings of the
  IEEE/CVF Conference on Computer Vision and Pattern Recognition (CVPR)}, 2020,
  pp. 1529--1538.

\bibitem{liu2018rethinking}
Z.~Liu, M.~Sun, T.~Zhou, G.~Huang, and T.~Darrell, ``Rethinking the value of
  network pruning,'' in \emph{Proceedings of the International Conference on
  Learning Representations (ICLR)}, 2019.

\bibitem{le2021network}
D.~H. Le and B.-S. Hua, ``Network pruning that matters: A case study on
  retraining variants,'' in \emph{Proceedings of the International Conference
  on Learning Representations (ICLR)}, 2021.

\bibitem{mao2017exploring}
H.~Mao, S.~Han, J.~Pool, W.~Li, X.~Liu, Y.~Wang, and W.~J. Dally, ``Exploring
  the granularity of sparsity in convolutional neural networks,'' in
  \emph{Proceedings of the IEEE/CVF Conference on Computer Vision and Pattern
  Recognition Workshop (CVPRW)}, 2017, pp. 13--20.

\bibitem{wortsman2019discovering}
M.~Wortsman, A.~Farhadi, and M.~Rastegari, ``Discovering neural wirings,'' in
  \emph{Proceedings of the Advances in Neural Information Processing Systems
  (NeurIPS)}, 2019, pp. 2684--2694.

\bibitem{xie2019exploring}
S.~Xie, A.~Kirillov, R.~Girshick, and K.~He, ``Exploring randomly wired neural
  networks for image recognition,'' in \emph{Proceedings of the IEEE/CVF
  International Conference on Computer Vision (ICCV)}, 2019, pp. 1284--1293.

\bibitem{howard2017mobilenets}
A.~G. Howard, M.~Zhu, B.~Chen, D.~Kalenichenko, W.~Wang, T.~Weyand,
  M.~Andreetto, and H.~Adam, ``Mobilenets: Efficient convolutional neural
  networks for mobile vision applications,'' \emph{arXiv preprint
  arXiv:1704.04861}, 2017.

\bibitem{sandler2018mobilenetv2}
M.~Sandler, A.~Howard, M.~Zhu, A.~Zhmoginov, and L.-C. Chen, ``Mobilenetv2:
  Inverted residuals and linear bottlenecks,'' in \emph{Proceedings of the
  IEEE/CVF Conference on Computer Vision and Pattern Recognition (CVPR)}, 2018,
  pp. 4510--4520.

\bibitem{howard2019searching}
A.~Howard, M.~Sandler, G.~Chu, L.-C. Chen, B.~Chen, M.~Tan, W.~Wang, Y.~Zhu,
  R.~Pang, V.~Vasudevan \emph{et~al.}, ``Searching for mobilenetv3,'' in
  \emph{Proceedings of the IEEE/CVF International Conference on Computer Vision
  (ICCV)}, 2019, pp. 1314--1324.

\bibitem{deng2009imagenet}
J.~Deng, W.~Dong, R.~Socher, L.-J. Li, K.~Li, and L.~Fei-Fei, ``Imagenet: A
  large-scale hierarchical image database,'' in \emph{Proceedings of the
  IEEE/CVF Conference on Computer Vision and Pattern Recognition (CVPR)}, 2009,
  pp. 248--255.

\bibitem{lecun1989optimal}
Y.~LeCun, J.~Denker, and S.~Solla, ``Optimal brain damage,'' in
  \emph{Proceedings of the Advances in Neural Information Processing Systems
  (NeurIPS)}, 1989, pp. 598--605.

\bibitem{hassibi1992second}
B.~Hassibi and D.~Stork, ``Second order derivatives for network pruning:
  Optimal brain surgeon,'' in \emph{Proceedings of the Advances in Neural
  Information Processing Systems (NeurIPS)}, 1992, pp. 164--171.

\bibitem{dong2017learning}
X.~Dong, S.~Chen, and S.~Pan, ``Learning to prune deep neural networks via
  layer-wise optimal brain surgeon,'' in \emph{Proceedings of the Advances in
  Neural Information Processing Systems (NeurIPS)}, 2017, pp. 4860--4874.

\bibitem{guo2016dynamic}
Y.~Guo, A.~Yao, and Y.~Chen, ``Dynamic network surgery for efficient dnns,'' in
  \emph{Proceedings of the Advances in Neural Information Processing Systems
  (NeurIPS)}, 2016, pp. 1379--1387.

\bibitem{frankle2019lottery}
J.~Frankle and M.~Carbin, ``The lottery ticket hypothesis: Finding sparse,
  trainable neural networks,'' in \emph{Proceedings of the International
  Conference on Learning Representations (ICLR)}, 2019.

\bibitem{lin2020dynamic}
T.~Lin, S.~U. Stich, L.~Barba, D.~Dmitriev, and M.~Jaggi, ``Dynamic model
  pruning with feedback,'' in \emph{Proceedings of the International Conference
  on Learning Representations (ICLR)}, 2020.

\bibitem{he2018soft}
Y.~He, G.~Kang, X.~Dong, Y.~Fu, and Y.~Yang, ``Soft filter pruning for
  accelerating deep convolutional neural networks,'' in \emph{Proceedings of
  the International Joint Conference on Artificial Intelligence (IJCAI)}, 2018,
  pp. 2234--2240.

\bibitem{ding2019approximated}
X.~Ding, G.~Ding, Y.~Guo, J.~Han, and C.~Yan, ``Approximated oracle filter
  pruning for destructive cnn width optimization,'' in \emph{Proceedings of the
  International Conference on Machine Learning (ICML)}, 2019, pp. 1607--1616.

\bibitem{lin2020channel}
M.~Lin, R.~Ji, Y.~Zhang, B.~Zhang, Y.~Wu, and Y.~Tian, ``Channel pruning via
  automatic structure search,'' in \emph{Proceedings of the International Joint
  Conference on Artificial Intelligence (IJCAI)}, 2020, pp. 673--679.

\bibitem{he2018amc}
Y.~He, J.~Lin, Z.~Liu, H.~Wang, L.-J. Li, and S.~Han, ``Amc: Automl for model
  compression and acceleration on mobile devices,'' in \emph{Proceedings of the
  European Conference on Computer Vision (ECCV)}, 2018, pp. 784--800.

\bibitem{molchanov2017variational}
D.~Molchanov, A.~Ashukha, and D.~Vetrov, ``Variational dropout sparsifies deep
  neural networks,'' in \emph{Proceedings of the International Conference on
  Machine Learning (ICML)}, 2017, pp. 2498--2507.

\bibitem{pool2021channel}
J.~Pool and C.~Yu, ``Channel permutations for n: m sparsity,'' in
  \emph{Proceedings of the Advances in Neural Information Processing Systems
  (NeurIPS)}, 2021, pp. 13\,316--13\,327.

\bibitem{han2016deep}
S.~Han, H.~Mao, and W.~J. Dally, ``Deep compression: Compressing deep neural
  networks with pruning, trained quantization and huffman coding,'' in
  \emph{Proceedings of the International Conference on Learning Representations
  (ICLR)}, 2016.

\bibitem{shahnaz2011blocked}
R.~Shahnaz and A.~Usman, ``Blocked-based sparse matrix-vector multiplication on
  distributed memory parallel computers.'' \emph{The International Arab Journal
  of Information Technology (IAJIT)}, vol.~8, no.~2, pp. 130--136, 2011.

\bibitem{chen2018tvm}
T.~Chen, T.~Moreau, Z.~Jiang, L.~Zheng, E.~Yan, H.~Shen, M.~Cowan, L.~Wang,
  Y.~Hu, L.~Ceze \emph{et~al.}, ``Tvm: An automated end-to-end optimizing
  compiler for deep learning,'' in \emph{Symposium on Operating Systems Design
  and Implementation (OSDI)}, 2018, pp. 578--594.

\bibitem{zheng2020ansor}
L.~Zheng, C.~Jia, M.~Sun, Z.~Wu, C.~H. Yu, A.~Haj-Ali, Y.~Wang, J.~Yang,
  D.~Zhuo, K.~Sen \emph{et~al.}, ``Ansor: Generating high-performance tensor
  programs for deep learning,'' in \emph{Symposium on Operating Systems Design
  and Implementation (OSDI)}, 2020, pp. 863--879.

\bibitem{tflite2019}
``Google llc. tensorflow lite,'' \url{https://www.tensorflow.org/lite}, [Online
  Accessed, 2019].

\bibitem{ji2018tetris}
Y.~Ji, L.~Liang, L.~Deng, Y.~Zhang, Y.~Zhang, and Y.~Xie, ``Tetris:
  Tile-matching the tremendous irregular sparsity,'' in \emph{Proceedings of
  the Advances in Neural Information Processing Systems (NeurIPS)}, vol.~31,
  2018, pp. 4115--4125.

\bibitem{narang2017block}
S.~Narang, E.~Undersander, and G.~Diamos, ``Block-sparse recurrent neural
  networks,'' \emph{arXiv preprint arXiv:1711.02782}, 2017.

\bibitem{kalchbrenner2018efficient}
N.~Kalchbrenner, E.~Elsen, K.~Simonyan, S.~Noury, N.~Casagrande, E.~Lockhart,
  F.~Stimberg, A.~Oord, S.~Dieleman, and K.~Kavukcuoglu, ``Efficient neural
  audio synthesis,'' in \emph{Proceedings of the International Conference on
  Machine Learning (ICML)}, 2018, pp. 2410--2419.

\bibitem{elsen2020fast}
E.~Elsen, M.~Dukhan, T.~Gale, and K.~Simonyan, ``Fast sparse convnets,'' in
  \emph{Proceedings of the IEEE/CVF Conference on Computer Vision and Pattern
  Recognition (CVPR)}, 2020, pp. 14\,629--14\,638.

\bibitem{vooturi2018hierarchical}
D.~T. Vooturi, D.~Mudigere, and S.~Avancha, ``Hierarchical block sparse neural
  networks,'' \emph{arXiv preprint arXiv:1808.03420}, 2018.

\end{thebibliography}

\begin{IEEEbiography}[{\includegraphics[width=1in,height=1.25in,clip,keepaspectratio]{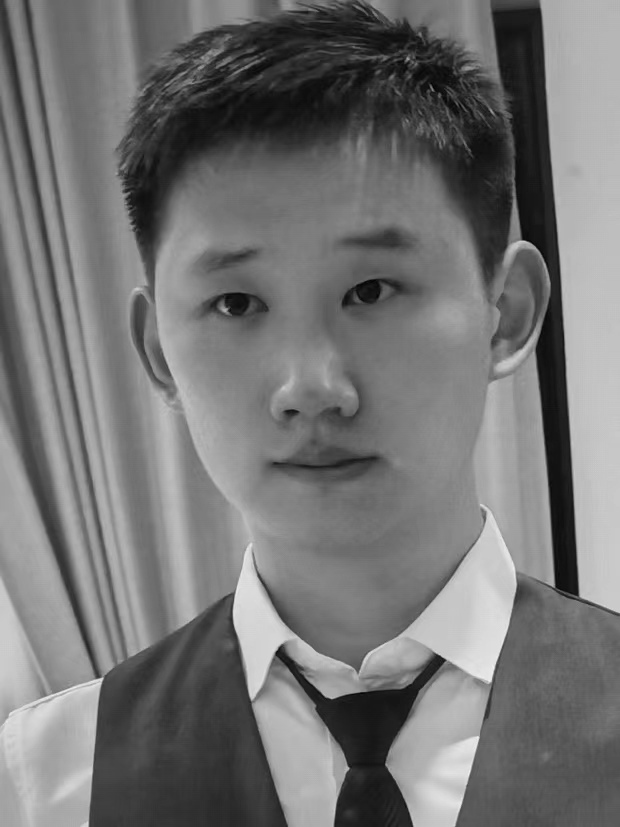}}]{Mingbao Lin} finished his M.S.-Ph.D. study and obtained the Ph.D. degree in intelligence science and technology from Xiamen University, Xiamen, China, in 2022. Earlier, he received the B.S. degree from Fuzhou University, Fuzhou, China, in 2016.

He is currently a senior researcher with the Tencent Youtu Lab, Shanghai, China. His publications on top-tier conferences/journals include IEEE TPAMI, IJCV, IEEE TIP, IEEE TNNLS, CVPR, NeurIPS, AAAI, IJCAI, ACM MM and so on. His current research interest is to develop efficient vision model, as well as information retrieval.
\end{IEEEbiography}

\begin{IEEEbiography}[{\includegraphics[width=1in,height=1.25in,clip,keepaspectratio]{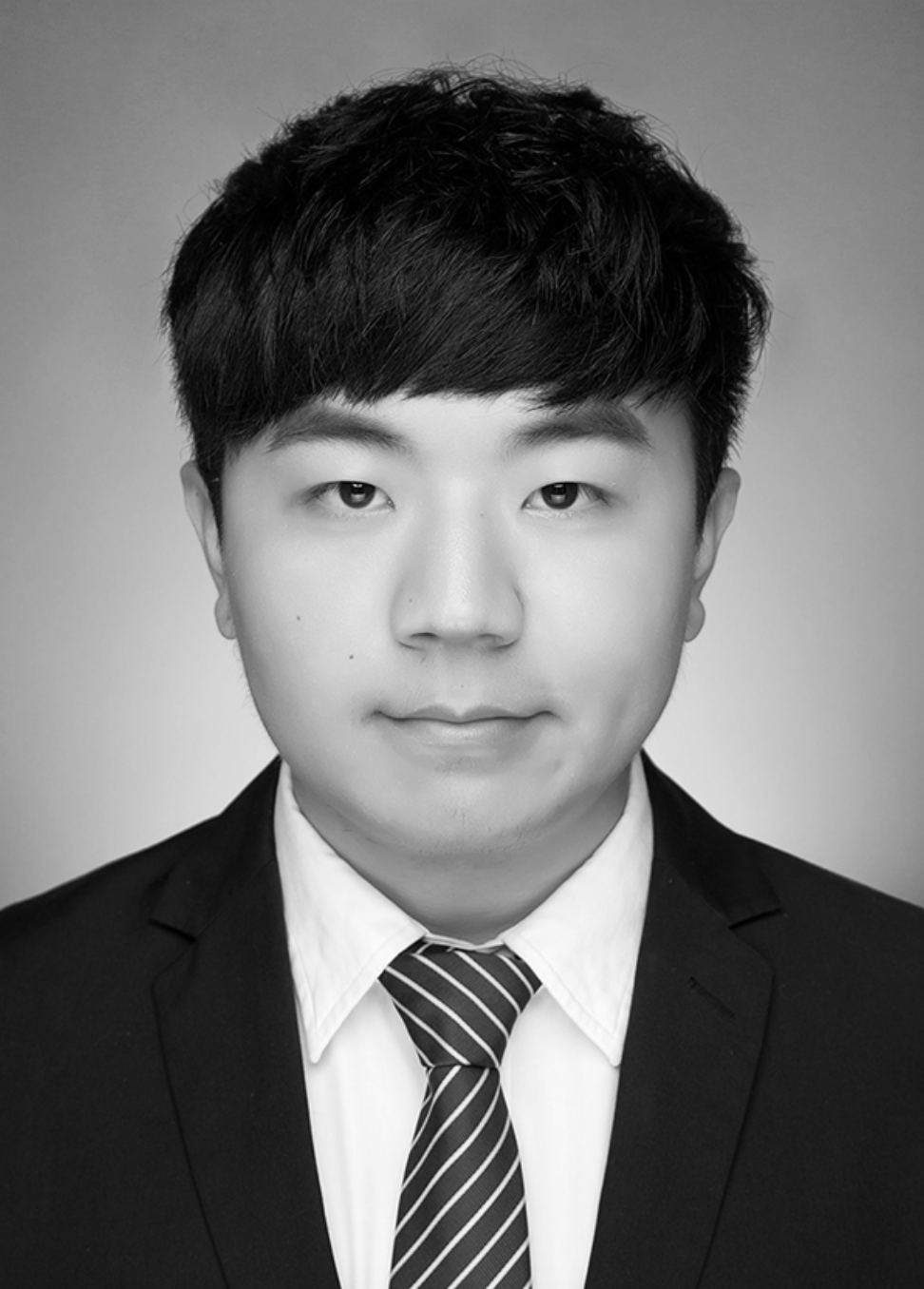}}]{Yuxin Zhang} received the B.E. degree in Computer Science, School of Informatics, Xiamen University, Xiamen, China, in 2020.
He is currently pursuing the B.S. degree with Xiamen University, China. His research interests include computer vision and neural network compression \& acceleration.
\end{IEEEbiography}

\begin{IEEEbiography}[{\includegraphics[width=1in,height=1.25in,clip,keepaspectratio]{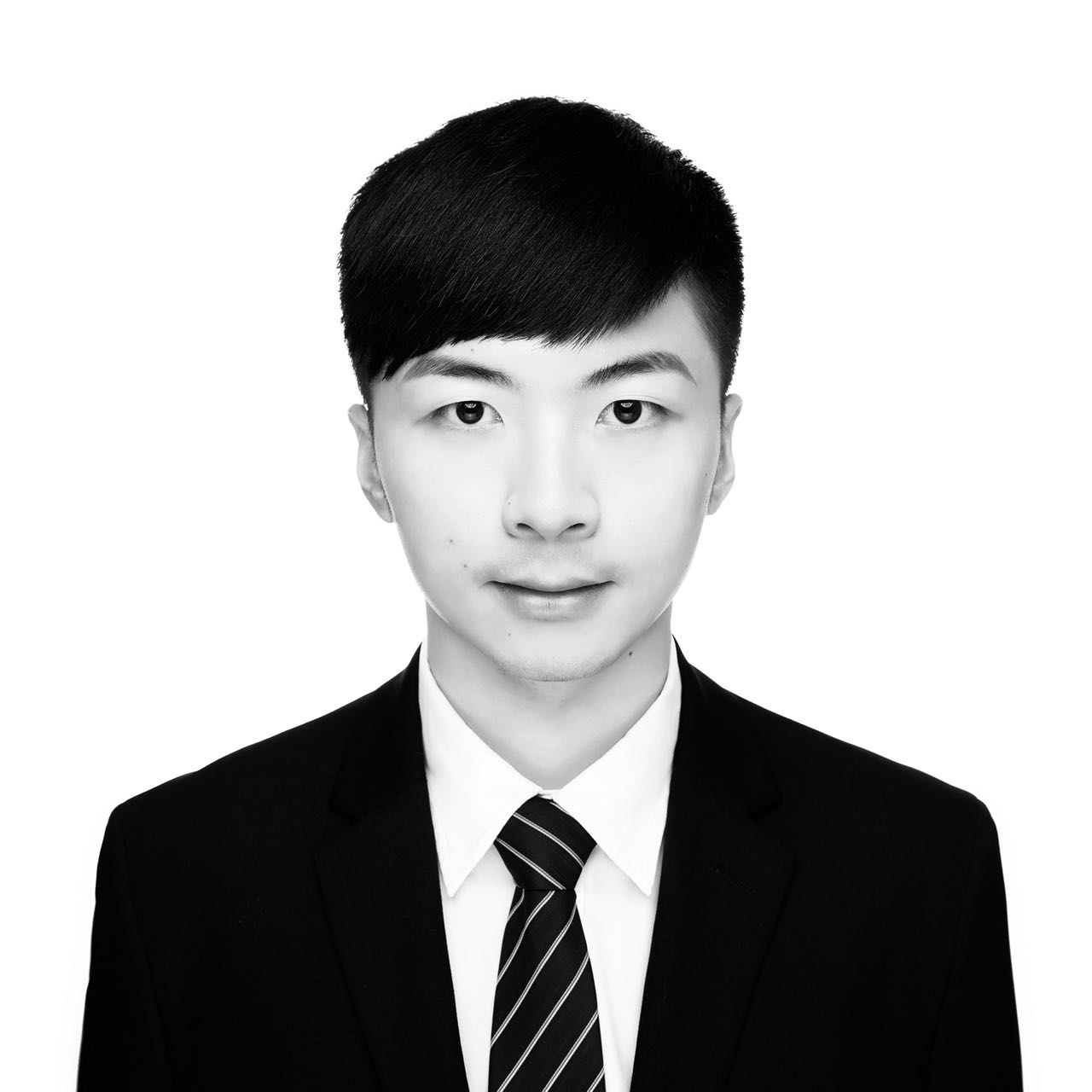}}]{Yuchao Li} received the M.S. degree in Computer Science, School of Information Science and Engineering, Xiamen University, Xiamen, China, in 2020. He is currently working toward the algorithm engineer in Alibaba. His research interests include computer vision and neural network compression and acceleration.
\end{IEEEbiography}

\begin{IEEEbiography}[{\includegraphics[width=1in,height=1.25in,clip,keepaspectratio]{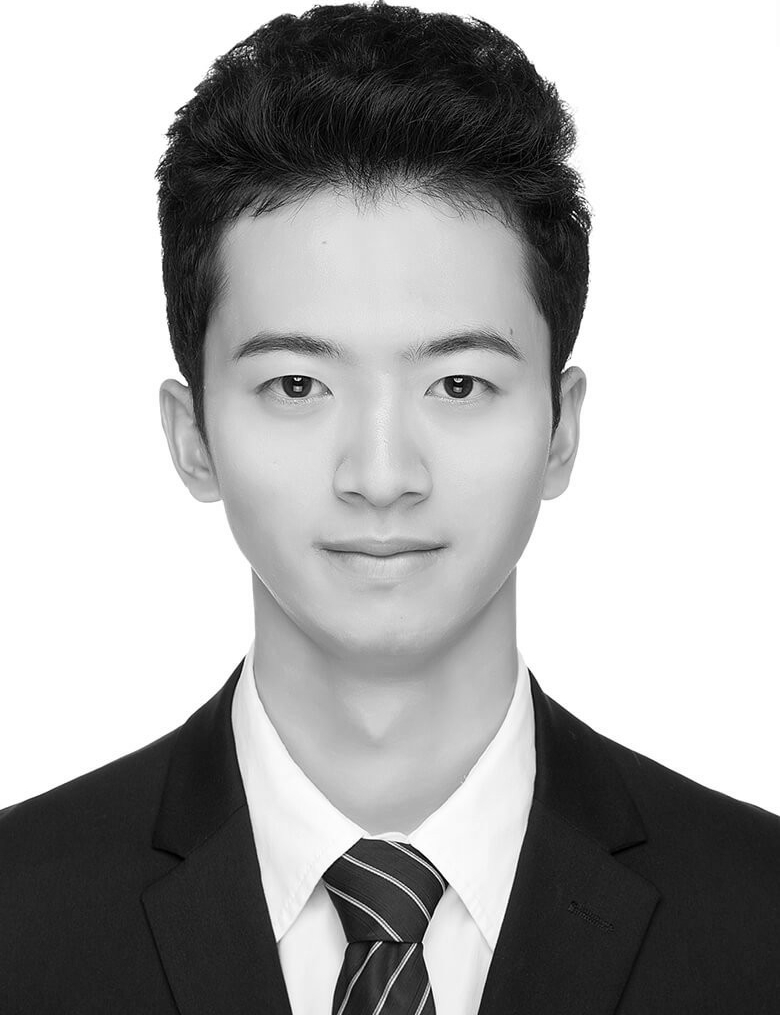}}]{Bohong Chen} received the B.E. degree in Computer Science, School of Informatics, Xiamen University, Xiamen, China, in 2020.
He is currently working toward the master’s degree from Xiamen University, China.
His research interests include computer vision, and neural network compression \& acceleration.
\end{IEEEbiography}

\begin{IEEEbiography}[{\includegraphics[width=1in,height=1.25in,clip,keepaspectratio]{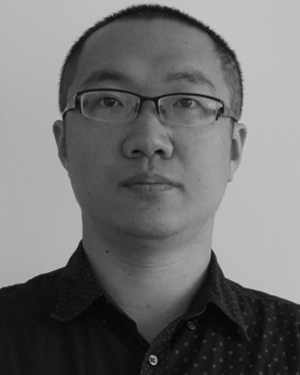}}]{Fei Chao}
(Member, IEEE) received the B.Sc. degree in mechanical engineering from the Fuzhou University, Fuzhou, China, in 2004, the M.Sc. degree with distinction in computer science from the University of Wales, Aberystwyth, U.K., in 2005, and the Ph.D. degree in robotics from the Aberystwyth University, Wales, U.K., in 2009.

He is currently an Associate Professor with the School of Informatics, Xiamen University, Xiamen, China. He has authored/co-authored more than 50 peer-reviewed journal and conference papers. His current research interests include developmental robotics, machine learning, and optimization algorithms.
\end{IEEEbiography}

\begin{IEEEbiography}[{\includegraphics[width=1in,height=1.25in,clip,keepaspectratio]{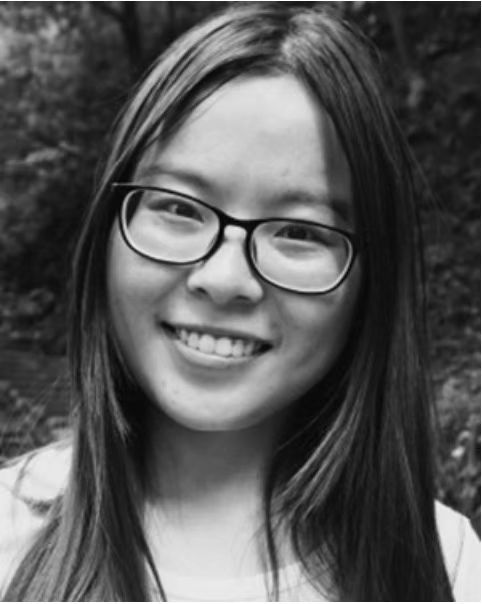}}]{Mengdi Wang} received the Ph.D. degree in Electronic Engineering, from Tsinghua University, Beijing, China, in 2017. She is currently working in Alibaba Group. Her research interests include efficient deep learning computing, model compression and neural architecture search.
\end{IEEEbiography}

\begin{IEEEbiography}[{\includegraphics[width=1in,height=1.25in,clip,keepaspectratio]{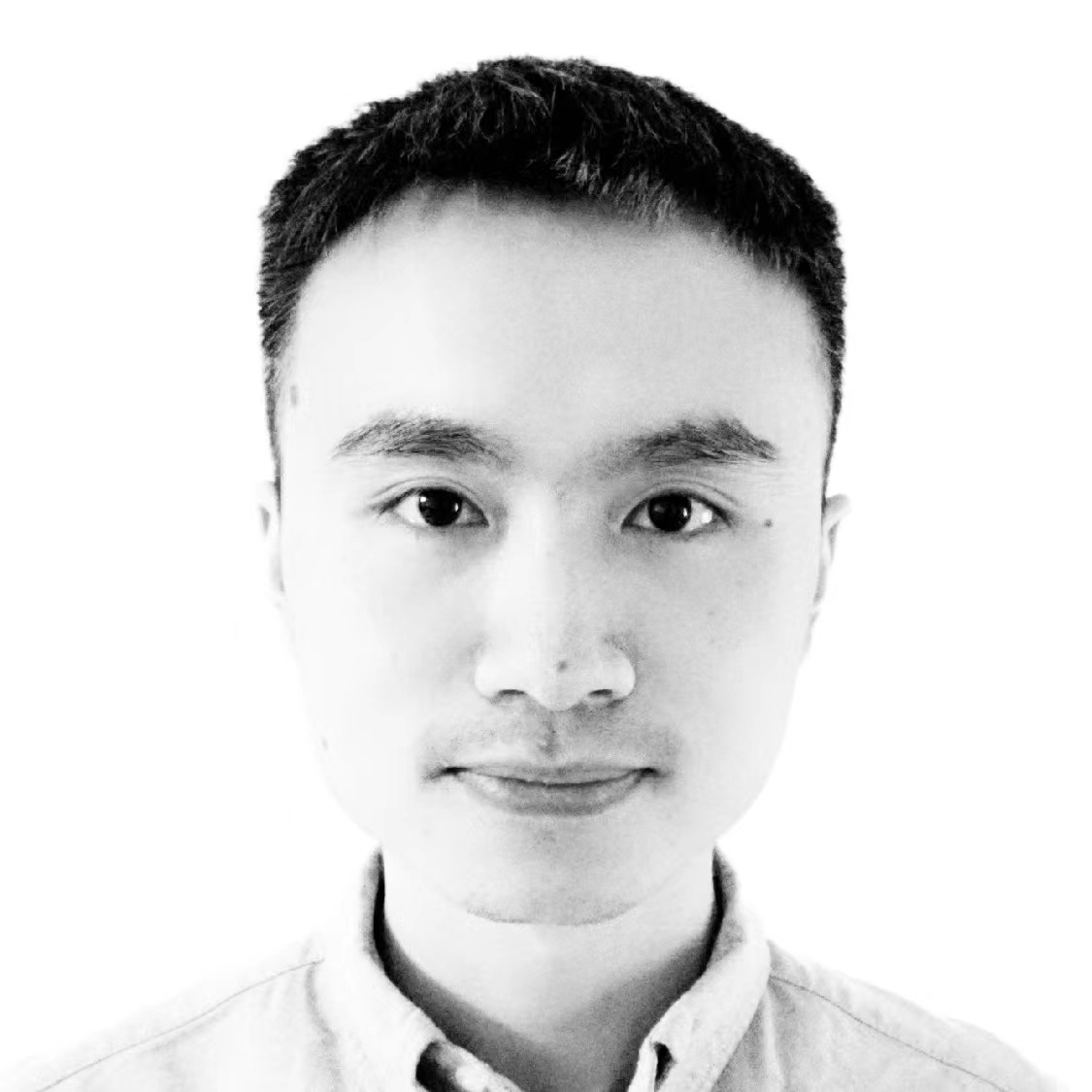}}]{Shen Li}  received his M.S. degree in Control Science and Engineering, Zhejiang University, Hangzhou, China, in 2013. He is currently working as Algorithm Expert at Platform of AI, Alibaba Cloud. His research interests include deep learning model compression and acceleration.
\end{IEEEbiography}

\begin{IEEEbiography}[{\includegraphics[width=1in,height=1.25in,clip,keepaspectratio]{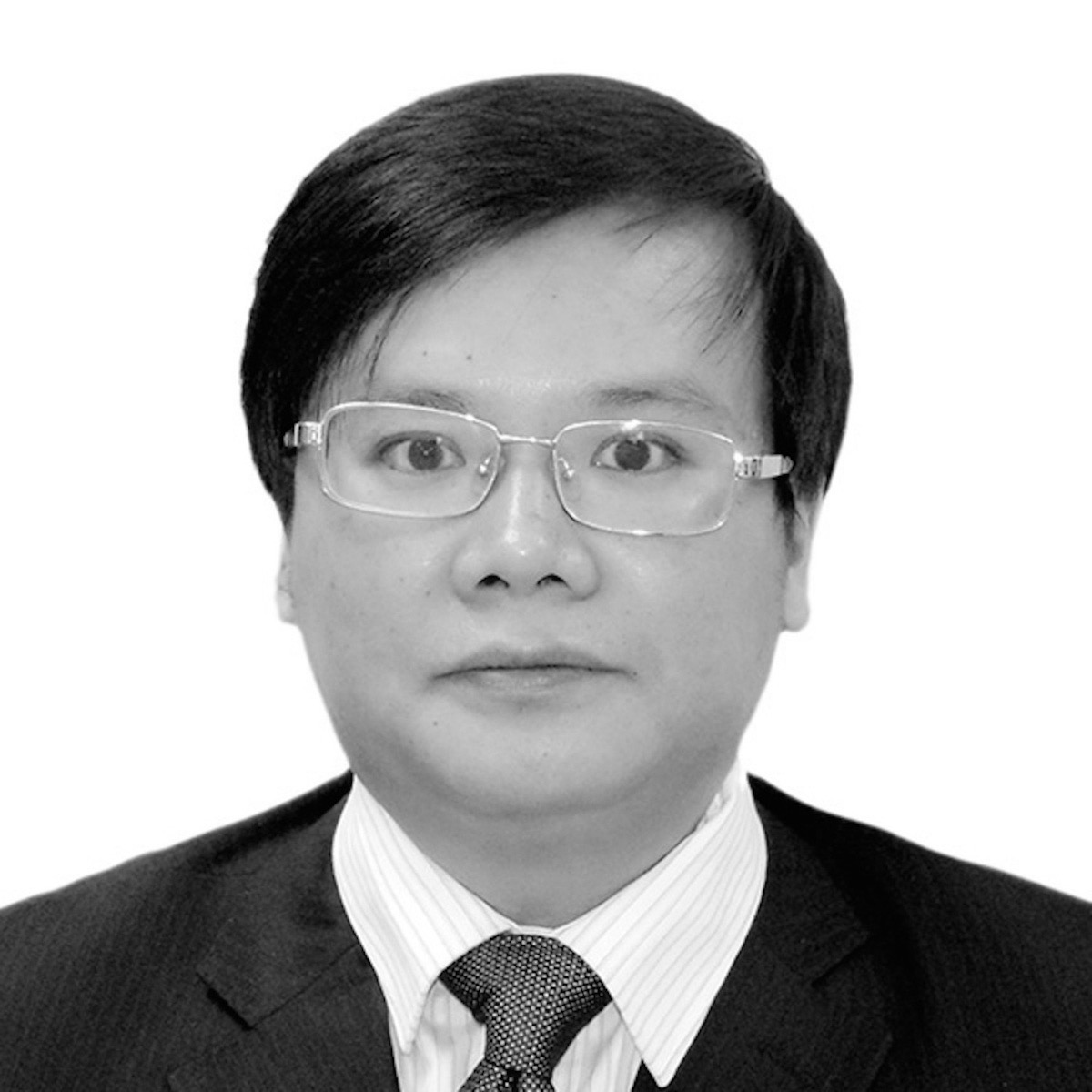}}]{Yonghong Tian} (Fellow, IEEE)
is currently a Boya Distinguished Professor with the Department of Computer Science and Technology, Peking University, China. His research interests include neuromorphic vision, brain-inspired computation and multimedia big data. He is the author or coauthor of over 200 technical articles in refereed journals such as IEEE TPAMI/TNNLS/TIP/TMM/TCSVT/TKDE/TPDS, ACM CSUR/TOIS/TOMM and conferences such as NeurIPS/CVPR/ICCV/AAAI/ACMMM/WWW. Prof. Tian was/is an Associate Editor of IEEE TCSVT (2018.1-), IEEE TMM (2014.8-2018.8), IEEE Multimedia Mag. (2018.1-), and IEEE Access (2017.1-). He co-initiated IEEE Int’l Conf. on Multimedia Big Data (BigMM) and served as the TPC Co-chair of BigMM 2015, and aslo served as the Technical Program Co-chair of IEEE ICME 2015, IEEE ISM 2015 and IEEE MIPR 2018/2019, and General Co-chair of IEEE MIPR 2020 and ICME2021. He is the steering member of IEEE ICME (2018-) and IEEE BigMM (2015-), and is a TPC Member of more than ten conferences such as CVPR, ICCV, ACM KDD, AAAI, ACM MM and ECCV. He was the recipient of the Chinese National Science Foundation for Distinguished Young Scholars in 2018, two National Science and Technology Awards and three ministerial-level awards in China, and obtained the 2015 EURASIP Best Paper Award for Journal on Image and Video Processing, and the best paper award of IEEE BigMM 2018. He is a senior member of IEEE, CIE and CCF, a member of ACM.
\end{IEEEbiography}

\begin{IEEEbiography}[{\includegraphics[width=1in,height=1.25in,clip,keepaspectratio]{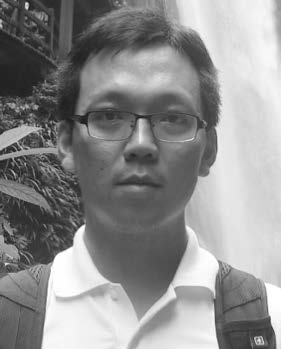}}]{Rongrong Ji}
(Senior Member, IEEE) is currently a Professor and the Director of the Intelligent Multimedia Technology Laboratory, and the Dean Assistant with the School of Information Science and Engineering, Xiamen University, Xiamen, China. His work mainly focuses on innovative technologies for multimedia signal processing, computer vision, and pattern recognition, with over 100 papers published in international journals and conferences. He is a member of the ACM. He was a recipient of the ACM Multimedia Best Paper Award and the Best Thesis Award of Harbin Institute of Technology. He serves as an Associate/Guest Editor for international journals and magazines such as \emph{Neurocomputing}, \emph{Signal Processing}, \emph{Multimedia Tools and Applications}, the \emph{IEEE Multimedia Magazine}, and the \emph{Multimedia Systems}. He also serves as program committee member for several Tier-$1$ international conferences.
\end{IEEEbiography}

\end{document}